%% file: main.tex
\documentclass{article}
\pdfoutput=1

\PassOptionsToPackage{numbers}{natbib}

    \usepackage[final]{neurips_2021}

\usepackage[utf8]{inputenc} 
\usepackage[T1]{fontenc}    
\usepackage{hyperref}       
\usepackage{url}            
\usepackage{booktabs}       
\usepackage{amsfonts}       
\usepackage{nicefrac}       
\usepackage{microtype}      
\usepackage{amsmath}
\usepackage{amssymb}
\usepackage{graphicx}
\usepackage{xcolor} 
\usepackage{wrapfig}
\usepackage{enumitem}

\title{Reinforcement Learning based Disease Progression Model for Alzheimer’s Disease}

\author{%
  Krishnakant V. ~Saboo \\
  UIUC\\
  \texttt{ksaboo2@illinois.edu}\\
  \And
  Anirudh Choudhary \\
  UIUC \\
  \texttt{ac67@illinois.edu} \\
  \AND
  Yurui Cao \\
  UIUC \\
  \texttt{yuruic2@illinois.edu} \\
  \And
  Gregory A. Worrell \\
  Mayo Clinic \\
  \texttt{Worrell.Gregory@mayo.edu} \\
  \And
  David T. Jones \\
  Mayo Clinic \\
  \texttt{Jones.David@mayo.edu} \\
  \And
  Ravishankar K. Iyer \\
  UIUC \\
  \texttt{rkiyer@illinois.edu} \\
}

\begin{document}

\maketitle

\begin{abstract}
We model Alzheimer’s disease (AD) progression by combining differential equations (DEs) and reinforcement learning (RL) with domain knowledge. DEs  provide relationships between some, but not all, factors relevant to AD. We assume that the missing relationships must satisfy general criteria about the working of the brain, for e.g., maximizing cognition while minimizing the cost of supporting cognition. This allows us to extract the missing relationships by using RL to optimize an objective (reward) function that captures the above criteria. We use our model consisting of DEs (as a simulator) and the trained RL agent to predict individualized 10-year AD progression using baseline (year 0) features on synthetic and real data. The model was comparable or better at predicting 10-year cognition trajectories than state-of-the-art learning-based models. Our interpretable model demonstrated, and provided insights into, "recovery/compensatory" processes that mitigate the effect of AD, even though those processes were not explicitly encoded in the model. Our framework combines DEs with RL for modelling AD progression and has broad applicability for understanding other neurological disorders.
\end{abstract}

\input{introduction}
\input{model}

\input{experiments}

\input{results}
\input{related_work}

\input{conclusion}

\begin{ack}

This work was supported by the Mayo Clinic and Illinois Alliance Fellowship for Technology-based Healthcare Research and in part by NSF grants CNS-1337732, CNS-1624790, and CCF-2029049 and the Jump ARCHES endowment fund. We thank Saurabh Jha, Subho Banerjee, Frances Rigberg, and Prakruthi Burra for their valuable feedback.
\end{ack}


\bibliographystyle{unsrt}
\bibliography{references}

\newpage

\input{appendix}



\end{document}

%% file: introduction.tex
\section{Introduction}

\begin{wrapfigure}{r}{0.4\textwidth}
  \begin{center}
    \includegraphics[width=0.43\textwidth]{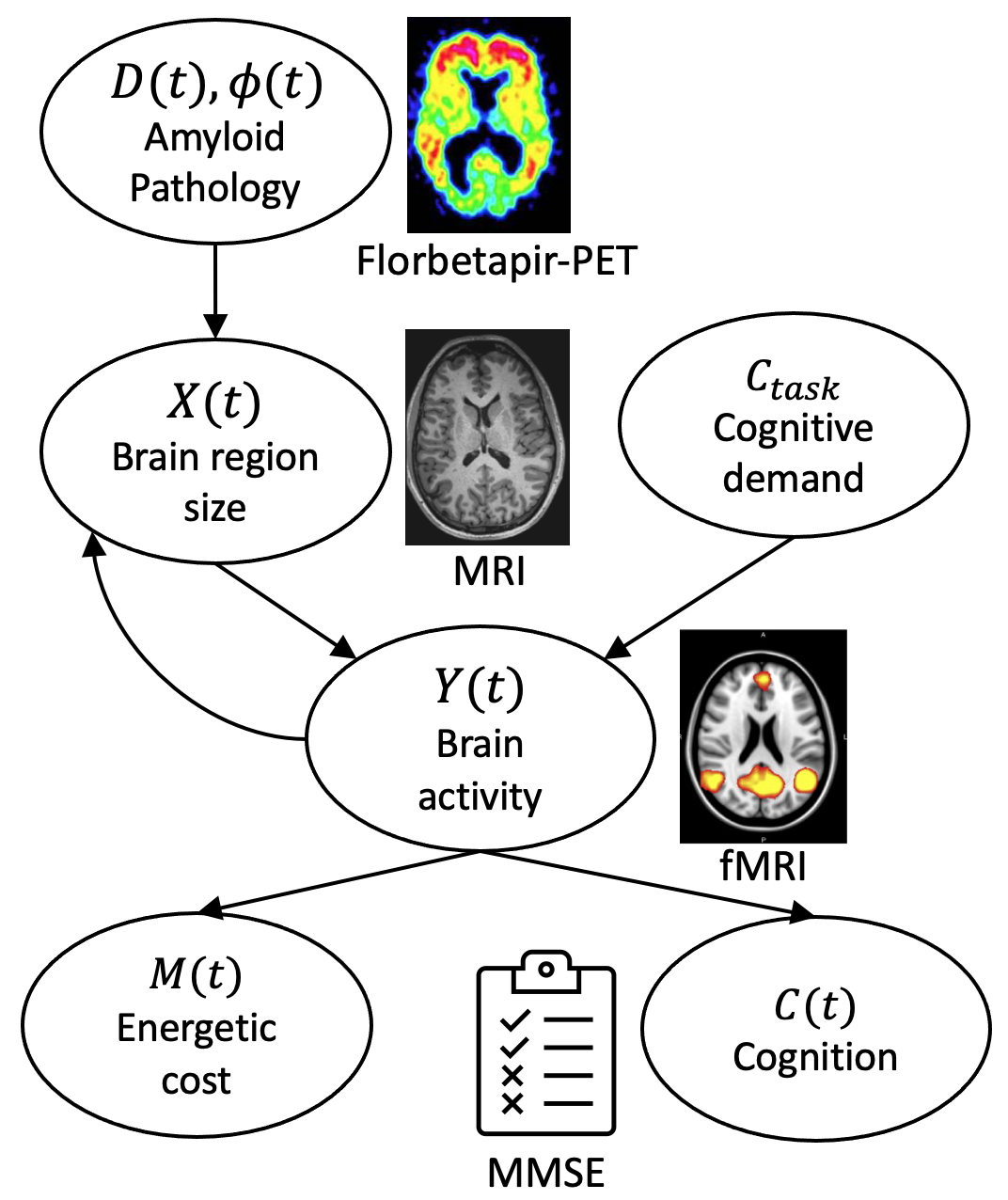}
  \end{center}
  \caption{Causal relationships between variables based on domain knowledge.\vspace{-1cm}}
  \label{fig:modelA}
\end{wrapfigure}

Models that describe Alzheimer's disease (AD) progression through time, i.e., the evolution of factors involved in the disease such as brain size, brain activity, pathology, and cognition (Fig. \ref{fig:modelA}), are crucial for mitigating this highly prevalent disease \cite{batsch2015world, breijyeh2020comprehensive}. Such models can enhance our \textit{understanding} of the disease processes and enable crucial applications like \textit{prediction} of long-term cognition trajectories for early detection. Our goal is to develop a model to predict an individual's future AD progression at 1-year intervals based on their baseline (year 0) data. We address this goal by combining differential equations that capture the relationships between some factors, and leveraging reinforcement learning to extract the missing relationships by optimizing a domain knowledge-based reward function. 

Differential\footnote{For simplicity, we use the term "differential equation" to denote algebraic and differential equations.} equation (DE) based models can describe disease progression by expressing domain knowledge as mathematical relationships between different factors \cite{raj2012network, weickenmeier2018multiphysics}. DE-based models have several advantages which make them attractive for disease progression modelling. The interpretability of these models can enhance our understanding of disease processes \cite{raj2015network, conrado2020challenges}. These models require limited data because the data is only used for parameter estimation. DE-based models' mechanistic nature allows for intervention exploration \cite{conrado2020challenges}. However, these models provide an incomplete view of the disease because DEs describing relationships among some of the factors are unavailable \cite{conrado2020challenges}.




We propose a framework for modeling AD progression that combines differential equations with reinforcement learning (RL) to overcome the above limitation. We assume that the missing relationships are the solution of an optimization problem that can be formulated based on domain knowledge. In our model, this optimization problem's objective function serves as the reward function for reinforcement learning. Therefore, by optimizing the reward, RL extracts the relationships among factors for which explicit DEs are unavailable. Thus, RL combined with the DEs describes the evolution of factors pertinent for modeling AD progression.





In our model, the available DEs define the simulator and the RL agent optimizes the domain-based reward function in the simulator. The parameters of the DEs are based on the available data. We overcome three main challenges for developing the proposed model. The first challenge is to identify the factors involved in each DE. We use domain knowledge to find existing causal relationships between different factors and differential equations relating them (when available).

Second, the DEs relating cognition, brain regions' sizes, and brain activity are unknown (Fig. \ref{fig:modelA}). Since proposing novel DEs relating these factors requires significant scientific knowledge that is still in development, we address this challenge as follows. Multiple brain regions work together to produce cognition \cite{bassett2009cognitive, ito2017cognitive}. We assume that the working of multiple brain regions is governed by an optimization problem that maximizes cognition while minimizing the cost of cognition \cite{christie2015cognitive}. We represent this optimization problem's objective function in terms of cognition, brain size, and brain activity.

Finally, the above optimization problem needs to be solved for multiple time points for modeling disease progression with the solution at time $t$ affecting the solution at time $t^{'}>t$. Therefore, we use RL to optimize the above objective (reward) function.



Our contributions are:
\begin{enumerate}[leftmargin=*]
    \item We developed an Alzheimer's disease progression model based on domain knowledge that combines DEs and RL. To the best of our knowledge, this is the first attempt at using RL to develop a disease progression model. Our framework is generic and can be used with other DEs, or to provide a basis for modelling other neurological disorders.
    \item We applied the model for predicting individualized long-term (10-year) future cognition trajectories using baseline ($0^{\text{th}}$ year) features on synthetic and real data. On real data, our model reduced the prediction error by $\sim 10 \%$ than a state-of-the-art deep learning-based model \cite{NGUYEN2020117203} and produced more realistic trajectories than other benchmark models. Cognition trajectory prediction models are useful in a clinical setting to identify individuals at future risk of cognitive decline.
    \item Our interpretable model provided insight into how multiple brain regions together produce cognition during AD. Specifically, we observed "recovery/compensatory" processes that mitigate the effect of AD on cognition \cite{hillary2017injured}, which could not be observed with state-of-the-art models. Recovery processes were not explicitly encoded into the model and were an outcome of the reward function's form. Further investigation into the basis of recovery processes could guide the development of interventions.
\end{enumerate}


\subsection{Background}
Alzheimer's disease results from the interaction of pathology (such as amyloid \cite{jack2010hypothetical}) and recovery (compensation \cite{davis2008pasa, hillary2017injured}) processes. These processes jointly affect brain structure (the size of different regions), brain function (activity in different regions), and cognition \cite{dennis2014functional, jack2010hypothetical, hillary2017injured, jones2017tau}. Pathology leads to neurodegeneration i.e., reduction in brain size \cite{jack2010hypothetical, braak2000vulnerability}, and consequently leads to cognitive decline. Recovery processes compensate for the effect of neurodegeneration on cognition through modification of brain activity by employing other regions of the brain \cite{davis2008pasa} for cognition. Ironically, the compensatory processes can lead to neurodegeneration and accelerated cognitive decline in the long term \cite{jones2017tau, hillary2017injured}. An individual's demographics such as gender, education, genetic risk for disease, etc., also play a role in determining disease progression \cite{macaulay2020predictors}.

%% file: model.tex
\section{Model}
First, we use domain knowledge to find existing causal relationships between different factors in Sec \ref{sec:causal}. In Sec \ref{sec:des}, we define the notation for the factors and substitute their causal relationships with appropriate DEs. In Sec \ref{sec:rl}, we formulate an optimization problem to find missing relationships and solve it using RL. Thus, the model predicts disease progression via the interaction between the DEs (which constitute a simulator) and the action of the RL agent. Finally, we describe the training of the model in Sec. \ref{sec:training}.

\begin{figure}
    \centering
    \includegraphics[width=\linewidth]{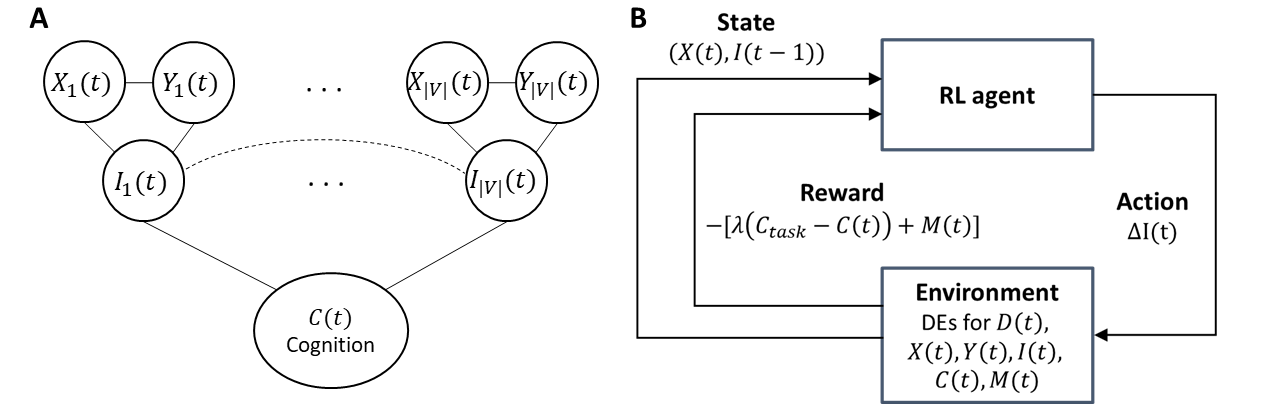}
    \caption{Framework for modeling AD progression. (A) Relationship between brain size, brain activity, information processing, and cognition (represented by solid and dashed edges). (B) Framework for AD progression that combines differential equations (DEs) with reinforcement learning (RL).}
    \label{fig:model}
\end{figure}

\subsection{Identification of causal interactions} \label{sec:causal}
We simplify the interactions between different factors by using the causal relationships between them proposed in literature (Fig. \ref{fig:modelA}). Amyloid beta (A$\beta$) (measured using florbetapir-PET) is a primary pathological factor in AD. It propagates from one brain region to another through tracts leading to its deposition in neighboring brain regions \cite{raj2012network}. Accumulation of amyloid directly affects brain structure (measured using MRI), i.e., brain regions sizes, and leads to neurodegeneration \cite{jack2010hypothetical}. Brain structure predicts brain activity (measured using fMRI)\cite{honey2007network, honey2009predicting}, and brain activity results in cognition (measured using cognitive tests for e.g., MMSE). Since brain activity also depends on cognitive task difficulty \cite{stern2012task}, we define a hypothetical variable, $C_{task}$, which represents the cognitive demand on the brain and can be thought of as the maximum score on a cognitive test. $C_{task}$ directly influences brain activity. Brain activity has an energetic cost, and itself can lead to neurodegeneration \cite{hillary2017injured}. Developing a model for AD progression is equivalent to describing the differential equation corresponding to each edge in Fig. \ref{fig:modelA}.


\subsection{Specification of differential equations} \label{sec:des}
\textbf{Brain structure:} We represent the brain as the graph $G_S=(V,E)$, where a node $v \in V$ represents a brain region, and an edge $e \in E$ represents a tract. Let $X_v(t)$ denote the size of a brain region $v \in V$ at time $t$, and $X(t) = [X_{1}(t), X_{2}(t),...,X_{|V|}(t)]$. 

\textbf{Pathology propagation:} To incorporate the propagation and accumulation of pathological A$\beta$ in our framework, we adapt the network diffusion based model proposed by Raj et al. \cite{raj2012network, raj2015network} because it captures the propagation of A$\beta$ through tracts. Let $D_v(t)$ be the instantaneous amyloid accumulation in region $v \in V$ at time $t$. Then propagation of amyloid is given by:
\begin{equation}
    \frac{dD(t)}{dt} = -\beta HD(t)
    \label{eq:amyloid_spread}
\end{equation}
where, $D(t)=[D_{1}(t), D_{2}(t), …, D_{|V|}(t)]$, $H$ is the Laplacian of the adjacency matrix of the graph $G_S$, and $\beta$ is a constant. The total amount of amyloid in a region, $\phi_v(t)$, is:
\begin{equation}
    \phi_v(t) = \int_{0}^{t} D_v(s)ds
    \label{eq:total_amyloid}
\end{equation}

\textbf{Brain activity and cognition:} Multiple brain regions work in synchrony to produce cognition \cite{ito2017cognitive}. Let $Y_v(t)$ denote the activity in region $v \in V$ in support of cognition $C(t)$ at time $t$. Although cognition, brain size ($X_v$), and activity ($Y_v$) are related, the exact relationship among them is unknown and cannot be easily learned from limited data. On the other hand, we can intuitively relate a region's size and activity to its "contribution" to cognition \cite{niven2007fly, davis2008pasa}. Therefore, we introduce a hypothetical variable termed as \textit{information processing}, $I_v(t)\in\mathbb{R}_{\geq0}$, to represent region $v$'s contribution to cognition (Fig. \ref{fig:model}). Therefore, the cognition $C(t)$ supported by the brain at time $t$ can be described as:
\begin{equation}
    C(t) = \sum_{v\in V} I_v(t).
    \label{eq:cog_info}
\end{equation}




The activity in a given region depends on the amount of information it is processing and its size \cite{niven2007fly, davis2008pasa}. For a region $v$ with size $X_v(t)$, the activity $Y_v(t)$ increases as $I_v(t)$ increases \cite{niven2007fly}. A healthier region (i.e., greater region size) can be considered more efficient and will require lesser activation than an inefficient one for comparable levels of information processing \cite{stern2012task}. Therefore, we propose the following relationship between $X_v(t), Y_v(t),$ and $I_v(t)$ with $\gamma$ as a constant:
\begin{equation}
    Y_v(t) = \gamma \frac{I_v(t)}{X_v(t)} \quad \forall v \in V
    \label{eq:energy_info_health}
\end{equation}
$I_v(t)$s are determined through reinforcement learning as described in Sec. \ref{sec:rl}.

\textbf{Energetic cost:} The brain consumes energy for supporting cognition. This energy consumption can be thought of as a cost to the brain. The energy consumption in a region is proportional to its activity \cite{attwell2001energy}. Therefore, the energetic cost to the brain is the total brain activity across all the regions. We compute the energetic cost $M(t)$ as follows:
\begin{equation}
    M(t) = \sum_{v \in V} Y_v(t)
    \label{eq:metacost_energy}
\end{equation}
\textbf{Degeneration of brain regions:} Neurodegeneration is influenced by two factors: amyloid deposition and brain activity. Previous studies that proposed a linear relationship between the rate of brain region degeneration and A$\beta$ deposition replicated the macroscopic neurodegeneration patterns seen in AD \cite{raj2015network, weickenmeier2018multiphysics}. Moreover, brain activity supporting cognition can further accelerate neurodegeneration \cite{hillary2017injured}. Therefore, we adopted the relationship proposed in \cite{raj2015network} and modified it to incorporate the effect of brain activity as follows:  
\begin{equation}
    \frac{dX_v(t)}{dt} = - \alpha_1 D_v(t) - \alpha_2 Y_v(t) \quad \forall v \in V
    \label{eq:health_atrophy}
\end{equation}
where $\alpha_1, \alpha_2$ are constants. The first term on the RHS in the Eq. \ref{eq:health_atrophy} is the degeneration due to A$\beta$ pathology and the second term is the deterioration due to brain activity (see Appendix \ref{sec:appex_brain_degen} for an alternate formulation of this equation).


\textbf{Parameter constants of equations:} An individual's demographics, such as gender, genetic risk of AD, education, etc., influence disease progression \cite{macaulay2020predictors}. In our model, the influence of demographics is mediated through the parameter constants $\alpha_1, \alpha_2, \beta, \gamma$ introduced in Eqs \ref{eq:amyloid_spread}, \ref{eq:energy_info_health}, and \ref{eq:health_atrophy} \cite{raj2015network}. For demographic factors $Z_0$ at baseline, let
\begin{equation}
    (\alpha_1, \alpha_2, \beta, \gamma) = f(Z_0).
    \label{eq:params_demogs}
\end{equation}

\subsection{Reinforcement learning to determine $I(t)$} \label{sec:rl}
Specifying the model requires determining the information processed by each region, i.e., computing $I(t) =[I_{1}(t),…,I_{|V|}(t)]$. Multiple brain regions jointly produce cognition. Those joint relationships influence $I(t)$. We assume that those joint relationships satisfy general criteria about the working of the brain and obtain $I(t)$ by solving an optimization problem capturing those criteria. Specifically, we assume that $I(t)$ is chosen to distribute the workload related to cognition "optimally" across multiple brain regions. The objective function of that optimization problem is based on two competing criteria: (i) minimizing the deficit between the cognitive demand of a task and the cognition $C(t)$ provided by the brain, and (ii) minimizing the energetic cost $M(t)$ of supporting cognition \cite{christie2015cognitive}. The optimization is performed at every time point $t$ with the solution at $t$ affecting the optimization problem at $t^{'}>t$. Therefore, we use RL to determine the optimal $I(t)$ by balancing the two competing criteria. RL is a desirable method in our case since it can train with limited data and also makes the framework generic and flexible to changes/refinements in the reward function and the underlying DEs. Below we describe how RL is adopted in our model (Fig. \ref{fig:model}). We implement the model for discrete time steps $t \in \{1,2,...,K\}$.

\textbf{State and action:} The state $S(t)$ consists of the current size of the brain regions $X(t)$ and the information processed by each region at the previous time point $I(t-1)$. The latter is provided in the state for ease of training the RL model. The action $A(t) \in \mathcal{A}$ specifies the change in information processed by each region from the previous time point, $\Delta I_v(t) \in \mathcal{R} \; \forall v \in V$. 
\begin{align}
\nonumber S(t) &= (X(t), I(t-1)), \nonumber\\
\mathcal{A} = \biggl\{ \left[\Delta I_{1}(t),…,\Delta I_{|V|}(t)\right] \biggr\rvert & I_v(t) = I_v(t-1)+\Delta I_v(t); \; \sum_{v\in V} I_v(t) \leq C_{task} \biggr\} \nonumber
\end{align}

\textbf{Environment:} The environment is a simulator consisting of the equations relating $D(t), \phi(t), X(t), Y(t), I(t), C(t)$, and $M(t)$ (Eqs. \ref{eq:amyloid_spread}, \ref{eq:total_amyloid}, \ref{eq:cog_info}, \ref{eq:energy_info_health}, \ref{eq:metacost_energy}, and \ref{eq:health_atrophy}). Based on the action $A(t)$, the environment updates the state and provides a reward to the RL agent.

\textbf{Reward:} The policy agent must balance the trade-off between the competing criteria of (i) reducing mismatch between $C_{task}$ and $C(t)$ and (ii) reducing the cost $M(t)$ of supporting cognition. Hence, we define the reward as follows:
\begin{equation}
    R(t) = -\left[ \lambda(C_{task} - C(t)) + M(t) \right]
    \label{eq:reward}
\end{equation}
where $\lambda$ is a parameter controlling the trade-off between the mismatch and the cost. The goal of the policy agent is to maximize the reward.

\subsection{Training}\label{sec:training}
\begin{figure}
    \centering
    \includegraphics[width=\linewidth]{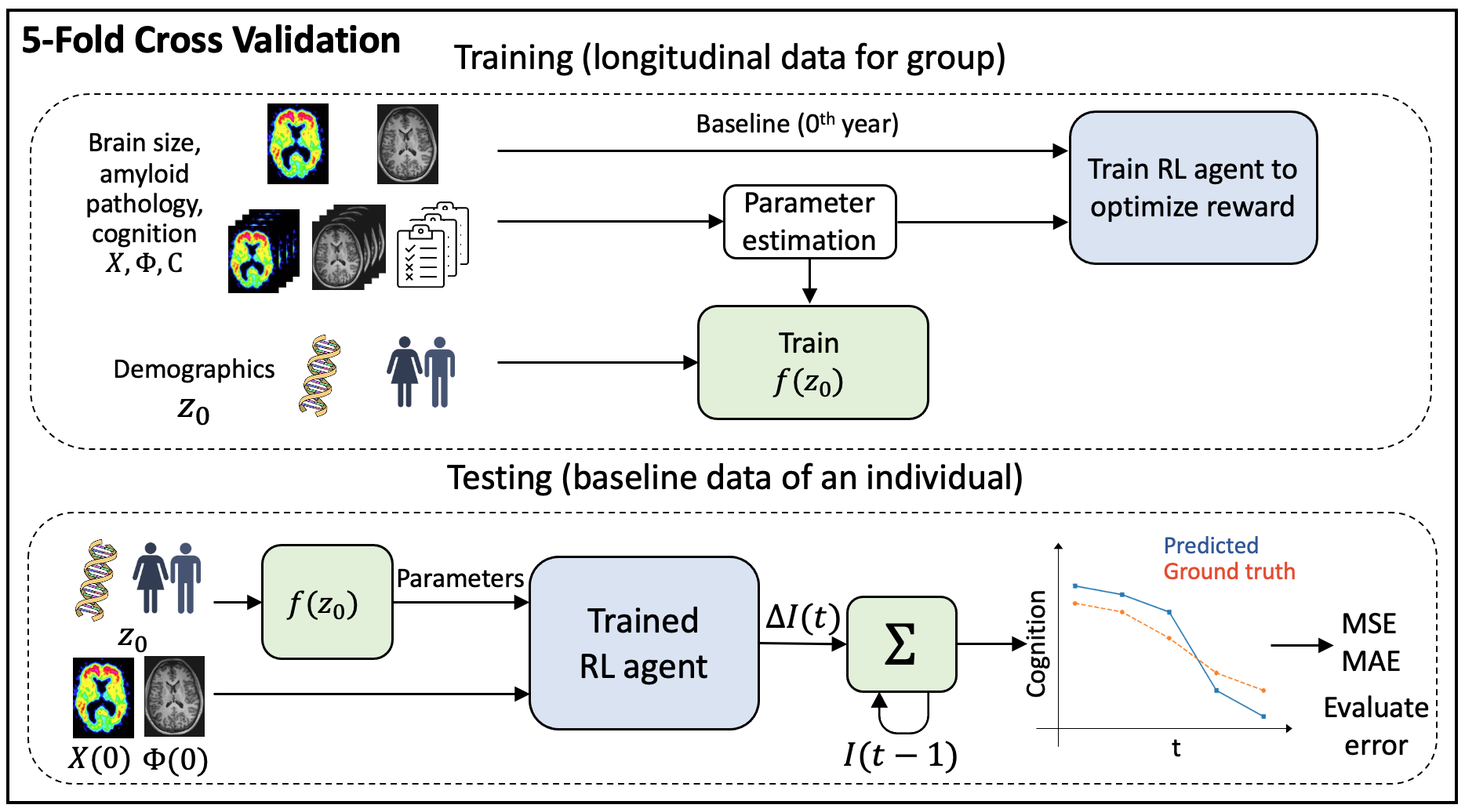}
    \caption{Workflow for training the proposed AD progression model and applying it for long-term cognition trajectory prediction.}
    \label{fig:traintest}
\end{figure}
Training the model consists of (i) estimating the parameters and finding $f$ in Eq. \ref{eq:params_demogs}, and (ii) training the RL model (Fig \ref{fig:traintest}). The model is trained using multimodal longitudinal data from individuals in the training set. First, we estimate the parameters $\alpha_1, \alpha_2, \beta, \gamma$ for people in the training set and relate the estimates to baseline demographic features of the individuals $Z_0$ to find $f$. Second, baseline measurements of the $X(0), \phi(0)$ are used with the estimated parameters to train the RL model.

\textbf{Parameter estimation:} We assume that for person $i \in \{1,...,N\}$, data at discrete time points $t\in \{1,...,K\}$ is available for the variables $X^{i}(t), D^{i}(t), Y^{i}(t), C^{i}(t)$. We derive parameter estimators by minimizing the L2 norm of the difference between the LHS and RHS of discretized versions of Eqs. \ref{eq:amyloid_spread} for $\beta$, Eq. \ref{eq:health_atrophy} for $\alpha_1, \alpha_2$, and Eqs. \ref{eq:cog_info} and \ref{eq:energy_info_health} for $\gamma$ (see Appendix \ref{sec:appex_param_es} for the derivations). 

Due to unavailability of functional MRI data in our experiments, we also derived parameter estimators for the case in which $Y(t)$ is unavailable for estimation. Minimizing the L2 norm obtained by eliminating $Y_{v}(t)$ from Eqs. \ref{eq:energy_info_health}, \ref{eq:health_atrophy}, and combining it with Eq. \ref{eq:cog_info} results in the following estimators (see Appendix \ref{sec:appex_param_es} for the derivation).
\begin{align}
    \hat{\beta} = -\frac{\sum_{i}\sum_{t}(D^{i}(t))^{T}H^{T} \frac{\Delta D^{i}(t)}{\Delta t}}{\sum_{i}\sum_{t}(D^{i}(t))^{T}H^{T}HD^{i}(t)}, \quad
    \hat{\alpha_1} = \frac{K_1 K_5 - K_3 K_4}{K_2 K_4 - K_3 K_5}, \quad
    \hat{\alpha_2 \gamma} = \frac{K_{3}^{2} - K_1 K_2}{K_2 K_4 - K_3 K_5}
    \label{eq:pares_alphas_gamma_withoutY}
\end{align}
where, $a^{i}_{1}(t) = (X^{i}(t))^{T}\frac{\Delta X^{i}(t)}{\Delta t}, a^{i}_{2}(t) = (X^{i}(t))^{T}D^{i}(t)$ are both scalars and $K_1 = \sum_{i}\sum_{t} (a^{i}_{1}(t))^{2}, K_2 = \sum_{i}\sum_t (a^{i}_{2}(t))^{2}, K_3 = \sum_{i}\sum_t a^{i}_{1}(t)a^{i}_{2}(t), K_4 = \sum_{i}\sum_t a^{i}_{1}(t)C^{i}(t), K_5 = \sum_{i}\sum_t a^{i}_{2}(t)C^{i}(t)$.

Note that in Eq. \ref{eq:pares_alphas_gamma_withoutY}, the product $\hat{\alpha_2 \gamma}$ cannot be resolved further due to unavailability of $Y(t)$. Therefore, we assumed that $\gamma$ is the same for all the individuals. Moreover, $\gamma$ affects the model either through $\alpha_2\gamma$ in Eq. \ref{eq:health_atrophy}, or essentially through $\lambda/\gamma$ in the reward function Eq. \ref{eq:reward} (see Appendix \ref{sec:appex_gamma_effect}). Therefore, we set $\gamma=1$ for all our experiments. 

Preliminary analysis showed that only discrete demographic features are associated with the parameter estimates (see Appendix \ref{sec:appex_param_es_adni}). Therefore, $f$ was the same as a table lookup based on $Z_0$. In a general setting, $f$ can be approximated with a linear model or a neural network.

\textbf{RL agent:} The agent is trained on the simulator with model-free on-policy learning using TRPO \cite{schulman2015trust} (see Appendix \ref{sec:appex_rltrain}). All the agent models are trained for 1 million episodes with a batch size of 1000. We clip the reward within a range of [-2000, 2000] with a KL-divergence of 0.01 and generalized advantage estimate discount factor of 0.97. We do not tweak any other hyperparameters of TRPO. 
The policy network is parameterized by a two-layer feedforward neural network with 32 hidden units each. We implement the simulator using OpenAI's Gym framework \cite{brockman2016openai}. We performed a grid-search for $\lambda$ and $I(0)$ and chose the value that minimized the validation set error (see Appendix \ref{sec:appex_grid}).


%% file: experiments.tex
\section{Experimental Setup}

\subsection{Adopting AD progression model for 10-year cognition trajectory prediction}
During testing, our model can be used for predicting the long-term future cognition trajectory of an individual based on their baseline ($0^{\text{th}}$ year) data (Fig. \ref{fig:traintest}). A new person's baseline demographics ($Z_0$) are used to find their parameters $\alpha_1, \alpha_2 \gamma, \beta$ using Eq. \ref{eq:params_demogs}. Those parameters, along with baseline value of brain regions sizes ($X(0)$) and amyloid ($\phi(0)$), are provided as initialization to the trained RL agent for predicting disease progression. We set cognitive demand $C_{task}=10$ for all the experiments. We predicted cognitive score at 1-year intervals for 10 years after baseline. 

\subsection{Dataset} 
We validated the model on synthetic data and real-world data derived from the Alzheimer's Disease Neuroimaging Initiative (ADNI) database (\href{http://adni.loni.usc.edu/}{adni.loni.usc.edu}) \cite{weiner2015impact}.

\textbf{Synthetic data:} We validated our model on synthetic data of 200 samples. Each sample consisted of longitudinal trajectories of the involved variables (see Appendix \ref{sec:appex_data} for details). 


\textbf{ADNI:} In this study, we included data from 160 individuals, used $|V|=2$ nodes representing the hippocampus (HC) and the prefrontal cortex (PFC), and measured cognition using Mini Mental State Examination (MMSE) (scaled by 3). We excluded fMRI scans from our analysis. See Appendix \ref{sec:appex_data} for details of data preprocessing. 

\subsection{Model evaluation} \label{sec:modeval}
Model evaluation consisted of (i) evaluating parameter estimation, (ii) validating group level AD progression trends, (iii) evaluating long-term cognition trajectory prediction, and (iv) providing insights into AD progression. We used 5-fold cross-validation for evaluation with a 64:16:20 split into training, validation, and testing sets. In each fold, each individual's data was only part of one of the three sets. Parameter estimators were evaluated only on the synthetic data using mean squared error, since ground truth values of parameters are known. For evaluating the model's predicted cognition trajectories, we compared them with ground truth trajectories using mean squared error (MSE) and mean absolute error (MAE). Since there are a large number of missing values in ADNI data, we only considered time points for which cognitive scores were available for an individual in computing prediction error. We compared the prediction performance with other state-of-the-art methods (\textbf{minimalRNN}, \textbf{SVR} \cite{NGUYEN2020117203}; see Appendix \ref{sec:appex_bench} for implementation details) for cognition trajectory prediction. To demonstrate the value of using RL in our framework, we compared the prediction with the case when $I(t)$ is determined by optimizing the reward $R(t)$ using grid-search separately at each time $t$ instead of RL (\textbf{w/o RL}) (see Appendix \ref{sec:appex_worl}). 






%% file: results.tex
\section{Results} \label{sec:results}
\begin{figure}
    \centering
    \includegraphics[width=\linewidth]{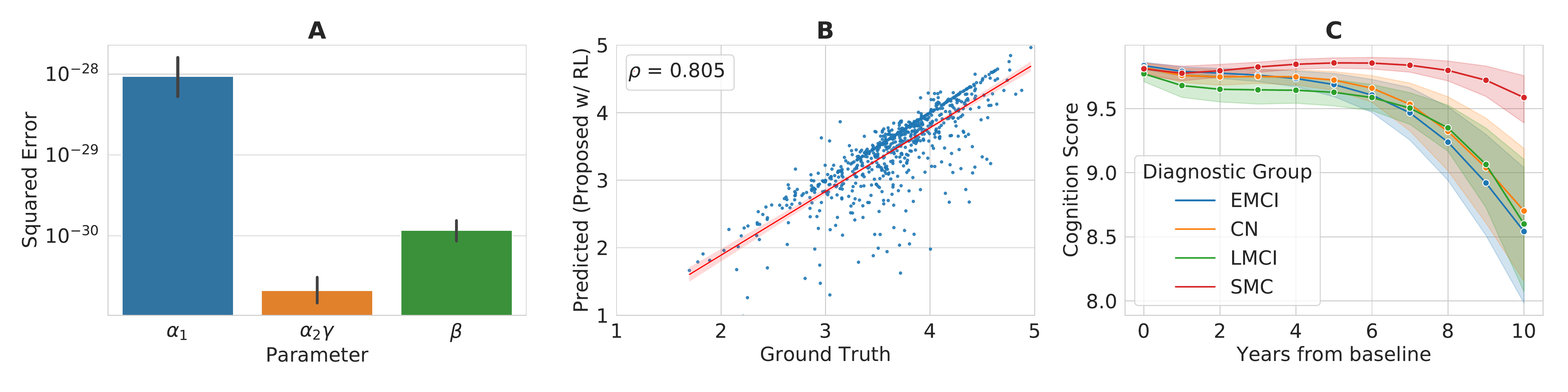}
    \caption{Validation of the proposed model. (A) Error in parameter estimation for synthetic data. Parameters were of the order of $10^{-2}$ to $10^{0}$. (B) Predicted vs. ground truth size of hippocampus for ADNI. (C) Average predicted cognition trajectories over 10 years for groups based on diagnosis at baseline in ADNI.}
    \label{fig:paraes}
\end{figure}
\subsection{Parameter estimation}
We estimated the parameters $\alpha_1, \alpha_2\gamma, \beta$ for the synthetic data using the estimators in Eq. \ref{eq:pares_alphas_gamma_withoutY}. Estimators achieved low errors (Fig. \ref{fig:paraes}). Since real data has a large number of missing values, we also evaluated the effect of missing values on parameter estimation. Although missing values increased the estimation error, the estimated parameters were still comparable to their ground truth values (see Appendix \ref{sec:appex_param_es_missing}). $Z_0$ consisted of two discrete variables for the synthetic data. Therefore, $f(Z_0)$ (from Eq. \ref{eq:params_demogs}) was a table lookup for the synthetic data.

For ADNI data, we performed preliminary analysis with individual-specific parameter estimates and chose gender and genetic risk (presence/absence of APOE-e4 genotype) as the demographic variables corresponding to $Z_0$ (see Appendix \ref{sec:appex_param_es_adni}). We also observed those parameter estimates were highly variable across subjects, which could be due to the small number of data points per individual, missing data, or noise in the data. Therefore, we estimated parameters for groups of subjects based on their gender and genetic risk. The estimated parameters are shown in the Appendix \ref{sec:appex_param_es_adni}. $f(Z_0)$ was a table lookup with $Z_0 =$ (gender, genetic risk).

\subsection{Model validation}
We validated the proposed model by comparing group-level trends for individuals in ADNI data. First, we compared the size of the brain regions obtained from our model with their size in the ground truth data. The values of regions size obtained from the model were highly correlated with those in the ground truth data (correlation of 0.8 for HC and 0.62 for PFC) (Fig. \ref{fig:paraes}, Fig. \ref{fig:pfc_ground_predicted}). We also compared the average cognition trajectories for each diagnostic group based on a person's diagnosis at baseline. As expected, the cognition of early and late mild cognitive impairment (MCI) individuals declined faster than individuals with significant memory concern (SMC) for whom cognitive impairment is less severe in reality (Fig. \ref{fig:paraes}). Counter-intuitively, cognitively normal (CN) individuals showed a rate of decline higher than SMC group and similar to MCI groups. To understand this result, we compared the mean baseline region sizes and amyloid values in different groups. Mean baseline brain region sizes for the CN group were lower than SMC and more similar to MCI groups (mean for HC- CN: 3.81, EMCI: 3.77, LMCI: 3.60, SMC: 3.94; PFC- CN: 3.71, EMCI: 3.81, LMCI: 3.74, SMC: 3.75). Similarly, mean baseline amyloid for CN participants was higher than SMC for HC (CN: 1.32, EMCI: 1.29, LMCI: 1.29, SMC:1.28) although for the frontal areas was smaller than SMC (CN: 1.26, EMCI: 1.30, LMCI: 1.32, SMC: 1.33). Since the disease progression prediction largely depends on the baseline value of the features, these observations taken together suggest that the similarity of baseline features of CN to MCI groups than SMC leads the predicted cognitive trajectories for CN being similar to MCI groups and worse than SMC.

\subsection{Cognition trajectory prediction}
\begin{figure}
    \centering
    \includegraphics[width=\linewidth]{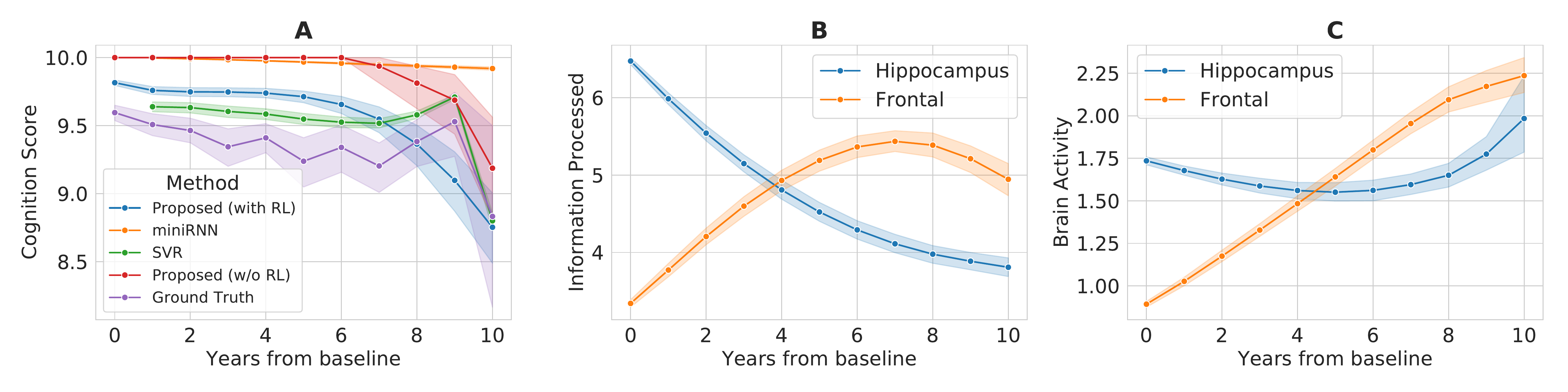}
    \caption{(A) Predicted 10-year future cognition trajectories on the ADNI averaged across individuals. Trajectories are shown for the ground truth, benchmark models, and proposed model. (B) Information processing averaged across individuals for HC (hippocampus) and PFC (frontal) during disease progression for the proposed model. (C) Brain activity averaged across individuals for HC and PFC during disease progression for the proposed model.}
    \label{fig:cogtraj}
\end{figure}

We evaluated the ability of the model to predict individualized 10-year future cognition trajectories based on baseline data (see Table \ref{tab:cogpred}). In general, data-driven models tend to be more accurate than mechanistic models and thus provide an upper limit for prediction performance (see Sec. \ref{sec:relatedwork}). The proposed framework achieved comparable or better performance than state-of-the-art methods on synthetic data as well as real data. The proposed model outperformed the DE-based model without RL, which demonstrates the value of using RL for combining DEs. At an individual level, predicted cognition from the model without RL was either 0 or 10 whereas predictions from the model with RL better resembled ground truth cognition (see Appendix \ref{sec:appex_worl}). The improved performance of the proposed model than the state-of-the-art deep learning-based (minimalRNN) model could be due to the small sample size. Although our model had a higher error compared to SVR on ADNI, the average cognition trajectories from our model were more realistic \cite{raket2020statistical} (Fig. \ref{fig:cogtraj}). Average trajectories for SVR on ADNI showed an increase in cognition at year 9 followed by a steep decline at year 10. This trend could be attributed to the disparity in the number of participants who had data at year 8 (n=60), year 9 (n=17), and year 10 (n=4) with few individuals having cognitive data at year 9 or 10. These observations highlight the value of using DEs and domain knowledge in our model. 

We also evaluated the consistency of the trajectories predicted by the proposed model when data from different follow-up times was used as "baseline". 134 participants had all the relevant data available at year 2. For those participants, we predicted trajectories using year 2 as "baseline" and compared them with the trajectories predicted using year 0 as baseline. The two trajectories for each participant were highly correlated (average correlation across participants=0.93).

\begin{table}[h]
    \centering
    \begin{tabular}{c c c c c}
        \hline
        \textbf{Method} & \multicolumn{2}{c}{\textbf{Synthetic Data}} & \multicolumn{2}{c}{\textbf{ADNI}} \\ \hline
        & MAE & MSE & MAE & MSE \\ \hline
        \textbf{Proposed (with RL)} & \textbf{0.641 (0.090)} & \textbf{0.910 (0.229)} & 0.537 (0.127) & 0.761 (0.370) \\
        Proposed (w/o RL) & 1.009 (1.670) & 3.806 (8.073) & 0.595 (0.137) & 1.112 (0.406) \\
        minimalRNN & 1.395 (0.149) & 6.971 (0.753) & 0.599 (0.137) & 0.984 (0.659) \\
        SVR & 0.658 (0.050) & 0.997 (0.112) & \textbf{0.495 (0.067)} & \textbf{0.574 (0.230)} \\\hline
    \end{tabular}
    \caption{Long-term cognition trajectory prediction performance. Standard deviations are provided in parentheses. Abbreviations: MAE, mean absolute error; MSE, mean squared error; and SVR, support vector regression.}
    \label{tab:cogpred}
\end{table}

\subsection{Demonstrating recovery}
We observed that our model exhibited recovery/compensatory processes during disease progression even though those were not explicitly encoded into the model. On averaging the trajectories of $I_v(t)$ for HC and PFC across individuals, we observed that as the contribution of HC to cognition reduced, the contribution of the PFC increased, resulting in maintaining the total cognition (Fig. \ref{fig:cogtraj}). As disease progressed further, the information processing in PFC also decreased while the information processing in HC continued to decrease, resulting in cognitive decline. This observation is consistent with recovery/compensatory behaviour that has been hypothesized to explain the response of the brain to injury \cite{davis2008pasa, hillary2017injured}. In control experiments, we observed that the proposed model trained with a modified reward function that only included the cognition mismatch term or the energetic cost term (in Eq. \ref{eq:reward}) did not show recovery processes (see Appendix \ref{sec:appex_control_mod}). Our observation of recovery processes emerging from the model formulation, specifically the reward function, is crucial because it could provide further understanding of the basis of those processes. Recovery-like behavior has also been seen in other neurological disorders \cite{hillary2017injured}, and understanding the mechanisms driving them could be useful for exploring interventions for AD. Our model also showed an initial phase of increased brain activity (hypermetabolism) in the PFC which is consistent with previous observations in AD and aging studies \cite{davis2008pasa, ashraf2015cortical}. The interpretable nature of our model enables these observations, which can help us better understand AD. Note that the interpretation of these results is meaningful within the context of the assumptions of the model and further research is needed to validate them.



%% file: related_work.tex
\section{Related Work} \label{sec:relatedwork}

Studies on modelling Alzheimer's disease progression can be broadly categorized as follows:
\begin{enumerate}
    \item \textit{Data-driven models} fit parametric/non-parametric models on biomarker data to relate disease pathology, region size, activity, cognition, and/or demographics. Examples of these models include event-based models \cite{fonteijn2012event}, mixed-effects models \cite{oxtoby2018data}, Bayesian models \cite{fruehwirt2018bayesian}, and machine learning-based approaches \cite{NGUYEN2020117203, lin2018convolutional, tabarestani2018profile, saboo2020predicting}. These methods have a low dependency on domain knowledge and work well for short-term prediction, but their long-term prediction performance is limited \cite{marinescu2019tadpole} due to lack of longitudinal multimodal data spanning the entire disease course. AD progresses over decades \cite{jack2011evidence, anderson2017so}, so accurate modeling requires long-term longitudinal data from multiple modalities (MRI, PET, cognitive assessments) \cite{jack2011evidence}, which is difficult to acquire. Further, interpretability of these models may be a challenge depending on their complexity which limits their applicability in understanding disease processes.
    
    \item \textit{Mechanistic models} use domain knowledge to represent the relationship among variables using algebraic and/or differential equations \cite{weickenmeier2018multiphysics, raj2012network}. Popular approaches include graph-based models \cite{li2016simulating, vertes2012simple} and dynamic causal modeling \cite{frassle2018generative}. These models are interpretable and have a low dependence on data. Although studies have modelled the relationships among a subset of the variables relevant for AD progression, to the best of our knowledge, a model that brings together pathology, brain structure, function, cognition, and demographics is yet to be proposed. 
\end{enumerate}

Our model enjoys the advantages offered by mechanistic models through the use of several DEs in its formulation. It differs from the existing mechanistic models by incorporating the relationship between brain size, activity, and cognition. Moreover, the use of an RL model trained in a DE-based simulator using a domain-guided reward function allows leveraging recent advances in machine learning without increasing the model's dependence on data.


Reinforcement learning has been used previously for developing treatment paradigms for neurological disorders such as epilepsy \cite{pineau2009treating} and Parkinson's disease \cite{krylov2020reinforcement}. Recently, Krylov et al. \cite{krylov2020reinforcement} showcased the efficacy of a policy-based RL framework in suppressing neuronal synchrony in Parkinson's disease. They incorporated DE-based neuronal models into a simulator and trained multiple PPO agents for oscillatory neuronal models. We are unaware of previous work that has used RL for AD disease progression modelling.

%% file: conclusion.tex
\section{Conclusion and Future Work} \label{sec:conclusion}
We proposed a framework for modeling AD progression by combining differential equations and reinforcement learning and evaluated the validity of its predictions on real and synthetic data. Our model successfully predicted individualized 10-year-long future cognition trajectories, which can be useful in a clinical setting for identifying individuals at a future risk of decline. Our model demonstrated, and provided insights into the potential basis of, recovery processes during AD progression which emerged from the model formulation, specifically the reward function. Thus, the model can help in further understanding disease progression. 
The mechanistic nature of our model allows the exploration of interventions strategies through perturbation analysis of the variables \cite{conrado2020challenges}. We contend that our framework is generic and can be modified to support other DEs describing AD, or provide a basis for modelling other neurological disorders because of the pervasiveness of recovery processes \cite{hillary2017injured} and the prevalence of using differential equations for modelling diseases \cite{venuto2016review, orlowski2013modelling}.

There are some limitations of the current work. We demonstrated the model with a 2-node graph, which is too coarse to capture the multimodal changes happening at finer spatial scales in the brain's structure and activity during AD progression. The model can scale to larger graphs with physiologically relevant nodes using data from the relevant brain regions. For relating brain activity to size, we proposed an inverse linear model (Eq. \ref{eq:energy_info_health}), but due to lack of sufficient fMRI data in our experiments, we could not validate that equation. Further experiments with fMRI data may lead to a better model for relating brain size ($X(t)$), activity ($Y(t)$), and information processing ($I(t)$). AD also results in tract degradation which we have not modeled. Related to this is the sensitivity of the model to misspecification of the dynamical system. We observed that the form of the DEs affects disease progression prediction (see Appendix \ref{sec:appex_inv_sq}) but the extent of the effect and its dependence on different factors needs further clarification. Finally, the model was only validated on ADNI data and, therefore, suffers from the same biases inherent in the data \cite{weiner2015impact}. Further experiments with larger and diverse data is needed to validate the results of this study. We plan to address these limitations in future work.




%% file: appendix.tex
\appendix
\section*{Appendix}
\section{Parameter estimators derivation and analyses} \label{sec:appex_param_es}
Parameters $\alpha_1, \alpha_2, \beta, \gamma$ introduced in Eqs. \ref{eq:amyloid_spread}, \ref{eq:health_atrophy}, \ref{eq:energy_info_health} are crucial to replicate the AD progression observed in real data. However, the appropriate values of the parameter are not known \textit{a priori} and have to be estimated from the data. Below, we derive optimal estimators for each of the parameters. For the following derivations, we assumed that longitudinal features $X^{i}(t), Y^{i}(t), \phi^{i}(t), D^{i}(t), C^{i}(t) \; \forall  \; t \in \{0, 1,2...,K\}$ are available for individual $i\in \{1,...,N\}$. Estimators are derived for a general setting of "same parameter for a group of individuals"  but the estimators are also applicable per-individual.

\subsection{Estimation of $\gamma$}
Parameter $\gamma$ in Eq. \ref{eq:energy_info_health} relates activity, size, and information processed by a region. Since $I(t)$ is not measured physically, we rearranged the equation to express it in terms of $\gamma, Y_v^{i}(t), X_v^{i}(t)$. We then used Eq. \ref{eq:cog_info} and setup an L2 minimization problem on the available data.

\begin{align*}
    \mathcal{L}\left(\frac{1}{\gamma}\right) &= \sum_{i}\sum_{t} \left(C^{i}(t) - \sum_{v\in V}\frac{1}{\gamma}Y_v^{i}(t)X_v^{i}(t) \right)^2 \\ 
    \gamma^* &= \arg\min_{\gamma \in \mathcal{R}} \mathcal{L}\left(\frac{1}{\gamma}\right)
\end{align*}
Replacing $\psi = \frac{1}{\gamma}$, and setting derivative to zero, we get:
\begin{align*}
    \frac{d\mathcal{L}(\psi)}{d\psi} &=  -\sum_{i}\sum_{t} 2\left(C^{i}(t) - \psi\sum_{v\in V}Y_v^{i}(t)X_v^{i}(t) \right)\sum_{v\in V}Y_v^{i}(t)X_v^{i}(t) \\
    0 &= \sum_{i}\sum_{t} C^{i}(t)(Y^{i}(t))^{T}X(t) - \psi\sum_{i}\sum_{t}((Y^{i}(t))^{T}X^{i}(t))^2 \\
    \psi &= \frac{\sum_{i}\sum_{t} C^{i}(t)(Y^{i}(t))^{T}X^{i}(t)}{\sum_{i}\sum_{t}((Y^{i}(t))^{T}X^{i}(t))^2}
\end{align*}

\subsection{Estimator for $\beta$}
$\beta$ can be estimated by setting up an L2 optimization problem using Eq. \ref{eq:amyloid_spread}. $D^{i}(t)$ and $H$ are available from measurements. We approximate $\frac{dD_v(t)}{dt} \approx \frac{\Delta D_v(t)}{\Delta t} \; \forall v \in V$, the RHS of which is available from longitudinal measurements of $D(t)$.
\begin{align*}
    \mathcal{L}(\beta) = \sum_{i}\sum_{t}\left\|\frac{\Delta D^{i}(t)}{\Delta t}+\beta H D^{i}(t)\right\|^{2}_{2}
    &= \sum_{i}\sum_{t}\left(\frac{\Delta D^{i}(t)}{\Delta t}+\beta H D^{i}(t)\right)^{T}\left(\frac{\Delta D^{i}(t)}{\Delta t}+\beta H D^{i}(t)\right)\\
    \beta^* &= \arg\min_{\beta \in \mathcal{R}}\mathcal{L}(\beta)
\end{align*}
Setting derivative wrt $\beta$ to zero, we get:
\begin{align*}
    \frac{d\mathcal{L}(\beta)}{d\beta} &= 2\sum_{i}\sum_{t} (HD^{i}(t))^{T}\left(\frac{\Delta D^{i}(t)}{\Delta t} + \beta H D^{i}(t) \right) \\
    0 &= \sum_{i}\sum_{t} (D^{i}(t))^{T}H^{T}\frac{\Delta D^{i}(t)}{\Delta t} + \beta \sum_{i}\sum_{t}(D^{i}(t))^{T}H^{T}HD^{i}(t) \\
    \beta &= - \frac{\sum_{i}\sum_{t}(D^{i}(t))^{T}H^{T}\frac{\Delta D^{i}(t)}{\Delta t}}{\sum_{i}\sum_{t}(D^{i}(t))^{T}H^{T}HD^{i}(t)}
\end{align*}

\subsection{Estimators for $\alpha_1, \alpha_2$}
$\alpha_1$ and $\alpha_2$ can be estimated using L2 norm optimization and Eq. \ref{eq:health_atrophy}. Rearranging terms in Eq. \ref{eq:health_atrophy}, we get:
\begin{align*}
    \frac{\Delta X^{i}_v(t)}{\Delta t} = -[D^{i}_v(t) \;\; Y^{i}_v(t)]\begin{bmatrix} \alpha_1 \\ \alpha_2\end{bmatrix} \quad \forall v \in V
\end{align*}
Stacking all the nodes together gives the following matrix notations, with $Q(t) = [D^{i}(t) Y^{i}(t)]$ is a matrix with $D(t)$ and $Y(t)$ as its columns, and $\alpha = [\alpha_1 \;\; \alpha_2]^{T}$,
\begin{align*}
    \frac{\Delta X^{i}(t)}{\Delta t} = -Q^{i}(t)\alpha
\end{align*}
The optimization problem is as follows:
\begin{align*}
    \mathcal{L}(\alpha) = \sum_{i}\sum_{t} \left \|\frac{\Delta X^{i}(t)}{\Delta t} +Q^{i}(t)\alpha \right\|_{2}^{2}
    &= \sum_{i}\sum_{t} \left (\frac{\Delta X^{i}(t)}{\Delta t} +Q^{i}(t)\alpha \right)^{T} \left (\frac{\Delta X^{i}(t)}{\Delta t} +Q^{i}(t)\alpha \right) \\
    \alpha^* &= \arg\min_{\alpha \in \mathcal{R}^2} \mathcal{L}(\alpha)
\end{align*}
Setting the derivative to 0 and simplifying, we get:
\begin{align*}
    \nabla \mathcal{L}(\alpha) &= \sum_{i}\sum_{t} 2(Q^{i}(t))^{T}\left (\frac{\Delta X^{i}(t)}{\Delta t} +Q^{i}(t)\alpha \right) \\
    0 &= \sum_{i}\sum_{t}(Q^{i}(t))^{T}\frac{\Delta X^{i}(t)}{\Delta t} + \sum_{i}\sum_{t}(Q^{i}(t))^{T}Q^{i}(t)\alpha \\
    \alpha &= -\left( \sum_{i}\sum_{t}(Q^{i}(t))^{T}Q^{i}(t) \right)^{-1} \left(\sum_{i}\sum_{t}(Q^{i}(t))^{T}\frac{\Delta X^{i}(t)}{\Delta t} \right)
\end{align*}

\subsection{Estimating parameters when $Y(t)$ is unavailable}
New parameter estimators that leverage only the available data need to be derived when $Y(t)$ is unavailable. The derivation goes as follows: first, we eliminate $Y(t)$ from the model equations. Second, we set up an L2 optimization problem involving the parameters and measured variables in the updated equation. Finally, we optimize the L2 objection function to derive optimal parameters.

Substituting Eq. \ref{eq:energy_info_health} in Eq. \ref{eq:health_atrophy} and expressing $I_v(t)$ in terms of the remaining variables, we get:
\begin{align*}
    \frac{dX_v(t)}{dt} &= -\alpha_1 D_v(t) -\alpha_2 \gamma \frac{I_v(t)}{X_v(t)} \\
    X_v(t)\frac{dX_v(t)}{dt} &= -\alpha_1 X_v(t)D_v(t) -\alpha_2 \gamma I_v(t) \\
    I_v(t) &= -\frac{1}{\alpha_2\gamma} \left (X_v(t)\frac{dX_v(t)}{dt} + \alpha_1 X_v(t)D_v(t) \right )
\end{align*}
Substituting the above expression in Eq. \ref{eq:cog_info}, denoting $\theta=(\alpha_1, \alpha_2\gamma)$, and setting up an L2 optimization problem based on it, we get:
\begin{align*}
    \mathcal{L}(\theta) &= \sum_{i,t} \biggl[C^{i}(t)
    + \frac{1}{\alpha_2\gamma}\sum_{v\in V} \left (X^{i}_v(t)\frac{dX^{i}_v(t)}{dt} + \alpha_1 X^{i}_v(t)D^{i}_v(t) \right ) \biggr]^{2}
\end{align*}

We approximate $\frac{dX^{i}_v(t)}{dt} \approx \frac{\Delta X^{i}_v(t)}{\Delta t}$. Additionally, $\sum_{v \in V} X^{i}_v(t)\frac{\Delta X^{i}_v(t)}{\Delta t} = (X^{i}(t))^{T}\frac{\Delta X^{i}(t)}{\Delta t}$  Similarly, $\sum_{v \in V} X^{i}_v(t) D^{i}_v(t) = (X^{i}(t))^{T}D^{i}(t)$. This gives:
\begin{align*}
    \mathcal{L}(\theta) &= \sum_{i,t} \biggl[C^{i}(t)  + \frac{1}{\alpha_2\gamma} \left ((X^{i}(t))^{T}\frac{\Delta X^{i}(t)}{\Delta t} + \alpha_1 (X^{i}(t))^{T}D^{i}(t) \right ) \biggr]^{2}
\end{align*}
To simplify notation, we replace $\delta_1 = \frac{1}{\alpha_2\gamma}$, $\delta_2 = \alpha_1$, $a^{i}_1(t)=(X^{i}(t))^{T}\frac{\Delta X^{i}(t)}{\Delta t}$, and $a^{i}_2(t)=(X^{i}(t))^{T}D^{i}(t)$.
\begin{align*}
    \mathcal{L}(\theta) &= \sum_{i,t} (C^{i}(t) + \delta_1 a^{i}_1(t) + \delta_1\delta_2 a^{i}_2(t) )^{2} \\
    \theta^* &= \arg\min_{\delta_1, \delta_2 \in \mathcal{R}} \mathcal{L}(\theta)
\end{align*}
$\delta_1$ and $\delta_2$ are coupled in the above optimization problem. We compute the partial derivative with respect to each variable and simply the resulting set of equations. First, computing the derivative with respect to $\delta_2$.
\begin{align}
    \frac{\partial \mathcal{L}(\theta)}{\partial \delta_2} &= \sum_{i,t}2(C^{i}(t) + \delta_1 a^{i}_1(t) + \delta_1 \delta_2 a^{i}_2(t))(0+0+\delta_1a_2(t)) \nonumber\\
    0 &= 2\delta_1 \sum_{i,t} a^{i}_2(t)(C^{i}(t) + \delta_1 a^{i}_1(t) + \delta_1 \delta_2 a^{i}_2(t)) \nonumber\\
    0 &= \sum_{i,t} a^{i}_2(t)C(t) + \delta_1 a^{i}_2(t)a^{i}_1(t) + \delta_1 \delta_2 (a^{i}_2(t))^{2} \nonumber\\
    \therefore\quad -\sum_{i,t} a_2(t)C(t) &= \delta_1 \sum_{i,t}a^{i}_2(t)a^{i}_1(t) + \delta_1 \delta_2 \sum_{i,t} (a^{i}_2(t))^{2} \nonumber \\ 
    \therefore\quad \delta_1 &= -\frac{\sum_{i,t} a^{i}_2(t)C^{i}(t)}{\sum_{i,t}a^{i}_2(t)a^{i}_1(t) + \delta_2 \sum_{i,t} (a^{i}_2(t))^{2}}
    \label{eq:delta2_in_delta1}
\end{align}
Now, computing partial derivative wrt $\delta_1$ and substituting its values from the Eq. \ref{eq:delta2_in_delta1}, we get
\begin{align}
    \frac{\partial \mathcal{L}(\theta)}{\partial \delta_1} &= \sum_{i,t} 2(C^{i}(t) + \delta_1 a^{i}_1(t) + \delta_1\delta_2 a^{i}_2(t)) ( a^{i}_1(t) + \delta_2 a^{i}_2(t)) \nonumber \\
    \therefore\quad 0 &= \sum_{i,t} C^{i}(t)(a^{i}_1(t) + \delta_2 a^{i}_2(t)) + \delta_1 (a^{i}_1(t) + \delta_2 a^{i}_2(t))^2 \nonumber \\
\therefore\quad \sum_{i,t} C^{i}(t)(a^{i}_1(t) + \delta_2 a^{i}_2(t)) &= -\sum_{i,t}\delta_1 (a^{i}_1(t) + \delta_2 a^{i}_2(t))^2 \nonumber\\
\therefore\quad \sum_{i,t} C^{i}(t)a^{i}_1(t) + \delta_2 \sum_{i,t}C^{i}(t)a^{i}_2(t) &= -\sum_{i,t}\delta_1 (a^{i}_1(t) + \delta_2 a^{i}_2(t))^2
\label{eq:partial_delta1}
\end{align}
Considering the RHS in the above equation separately for simplification, we get,
\begin{align*}
RHS &= -\sum_{i,t}\delta_1 ((a^{i}_1(t))^{2} + 2\delta_2 a^{i}_1(t)a^{i}_2(t) + \delta_2^{2} (a^{i}_2(t))^{2})\\
&= -\delta_1 \left(\sum_{i,t}a_1^{2}(t) + 2\delta_2 \sum_{i,t}a_1(t)a_2(t) + \delta_2^{2} \sum_{i,t}a_2^{2}(t)\right)
\end{align*}
We make the following substitutions for ease of notation: $K_1 = \sum_{i,t}(a^{i}_1(t))^{2}$, $K_2 = \sum_{i,t}(a^{i}_2(t))^{2}$, $K_3 = \sum_{i,t}a^{i}_1(t)a^{i}_2(t)$, $K_4 = \sum_{i,t}a^{i}_1(t)C^{i}(t)$, and $K_5 = \sum_{i,t}a^{i}_2(t)C^{i}(t)$. LHS and RHS of Eq. \ref{eq:partial_delta1} can be re-written with the above constants as follows:
\begin{align*}
    LHS &= \sum_{i,t} C^{i}(t)a^{i}_1(t) + \delta_2 \sum_{i,t}C^{i}(t)a^{i}_2(t)\\
    &= K_4 + \delta_2 K_5 \\
    RHS &= -\delta_1 \left(\sum_{i,t}a_1^{2}(t) + 2\delta_2 \sum_{i,t}a_1(t)a_2(t) + \delta_2^{2} \sum_{i,t}a_2^{2}(t)\right)\\
    &= -\delta_1 (K_1 + 2\delta_2 K_3 + \delta_2^{2} K_2) \\
    &= \left(\frac{K_5}{K_3 + \delta_2 K_2}\right) (K_1 + 2\delta_2 K_3 + \delta_2^{2} K_2)
\end{align*}

We equate LHS=RHS. That gives:
\begin{align*}
    (K_4 + \delta_2 K_5) (K_3 + \delta_2 K_2)
    &= K_5 (K_1 + 2\delta_2 K_3 + \delta_2^{2} K_2) \\
    \therefore\quad K_3K_4 + \delta_2K_2K_4 + \delta_2K_3K_5 +  \delta_2^{2}K_2K_5
    &= K_1 K_5 + 2\delta_2 K_3K_5 + \delta_2^{2} K_2K_5 \\
    \therefore\quad K_3K_4 + \delta_2K_2K_4 + \delta_2K_3K_5 
    &= K_1 K_5 + 2\delta_2 K_3K_5 \\
    \therefore\quad \delta_2K_2K_4 + \delta_2K_3K_5 -2\delta_2 K_3K_5
    &= K_1 K_5 - K_3K_4  \\
    \therefore\quad \delta_2(K_2K_4 + K_3K_5 -2K_3K_5)
    &= K_1 K_5 - K_3K_4  \\
    \delta_2 &= \frac{K_1 K_5 - K_3K_4}{K_2K_4 - K_3K_5}
\end{align*}
Substituting $\delta_2$ back in Eq. \ref{eq:delta2_in_delta1} and simplifying, we get:
\begin{align*}
    \delta_1 &= \frac{K_2K_4 - K_3 K_5}{K_3^{2} - K_1 K_2}
\end{align*}
Rewriting in terms of the parameters, 
\begin{align*}
    \alpha_1 &= \frac{K_1 K_5 - K_3K_4}{K_2K_4 - K_3K_5} \\
    \frac{1}{\alpha_2\gamma} &= \frac{K_2K_4 - K_3 K_5}{K_3^{2} - K_1 K_2}
\end{align*}

\subsection{Effect of missing values on parameter estimation} \label{sec:appex_param_es_missing}
\begin{wrapfigure}{r}{0.5\textwidth}
  \begin{center}
    \includegraphics[width=0.5\textwidth]{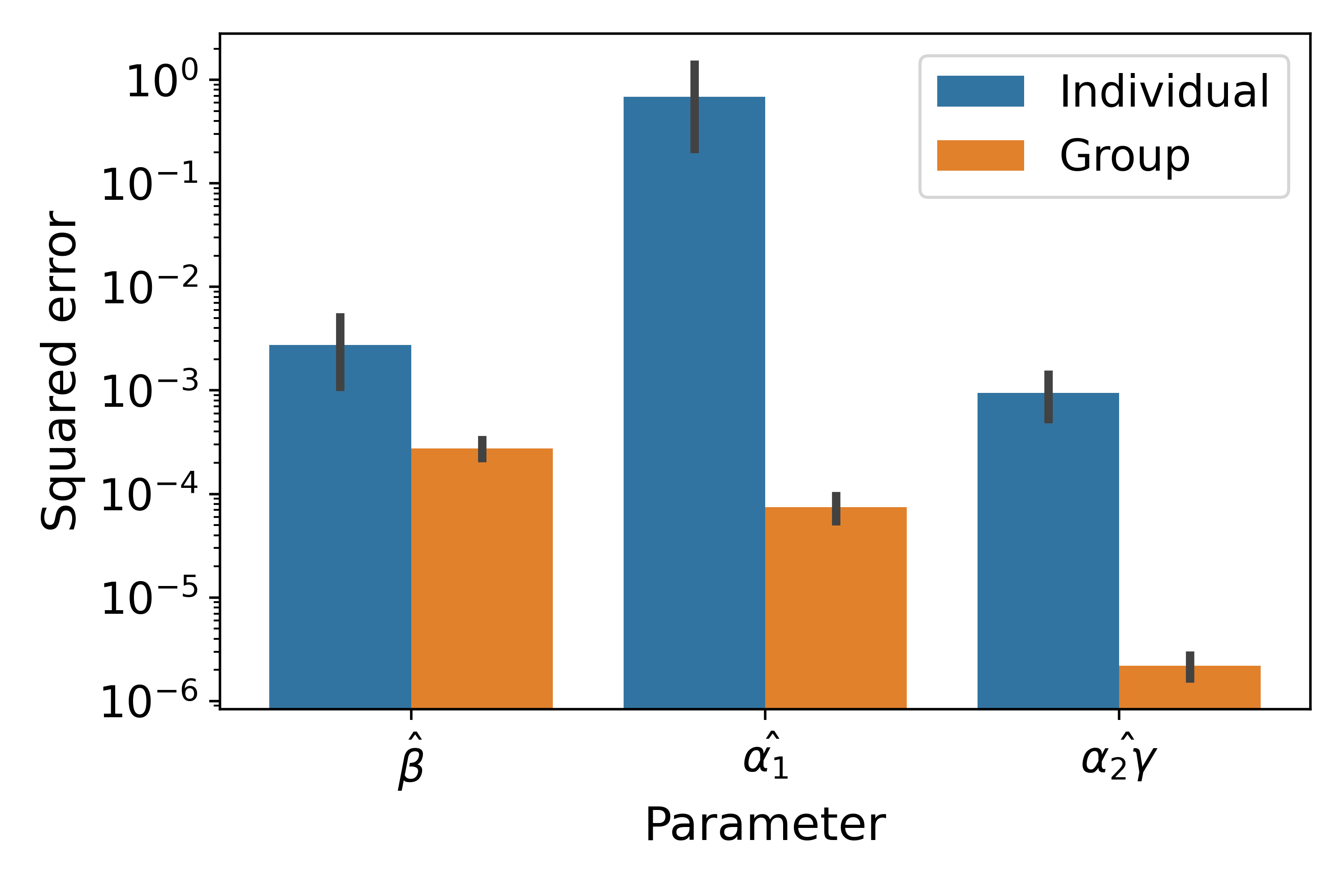}
  \end{center}
  \caption{Effect of missing values on parameter estimation. Analysis was performed for the synthetic data.}
  \label{fig:missing_paraes_error}
\end{wrapfigure}
In the ADNI data in this study, data is available only for $31\%$ of all possible visits (11 in total) for an individual on average -- there are a large number of missing values. To assess the effect of missing values on parameter estimation, we performed parameter estimation on the synthetic data by artificially removing measurements from its samples' trajectories. In the synthetic data, visits from a given year were randomly removed across the population so as to match the percentage of samples missing for that visit year in the real data. Data was removed in this way to ensure that the follow-up years data availability in the synthetic data matched the real data. On this synthetic data with missing values, we estimated the parameters for each individual separately and for a groups of individuals based on their demographics. The squared error of the estimated parameters are shown in Fig. \ref{fig:missing_paraes_error}. Although the estimation error was higher compared to the case when all the data was available (Fig. \ref{fig:paraes}), estimating parameters for group of individuals achieved lower error than individualized estimation. The ground truth parameter values are of the order of $10^{-2}$ to $10^{0}$. Therefore, estimation errors for groups are tolerable.

\subsection{ADNI - Selection of demographic variables and parameter estimates} \label{sec:appex_param_es_adni}
\begin{figure}
    \centering
    \includegraphics[width=0.95\linewidth]{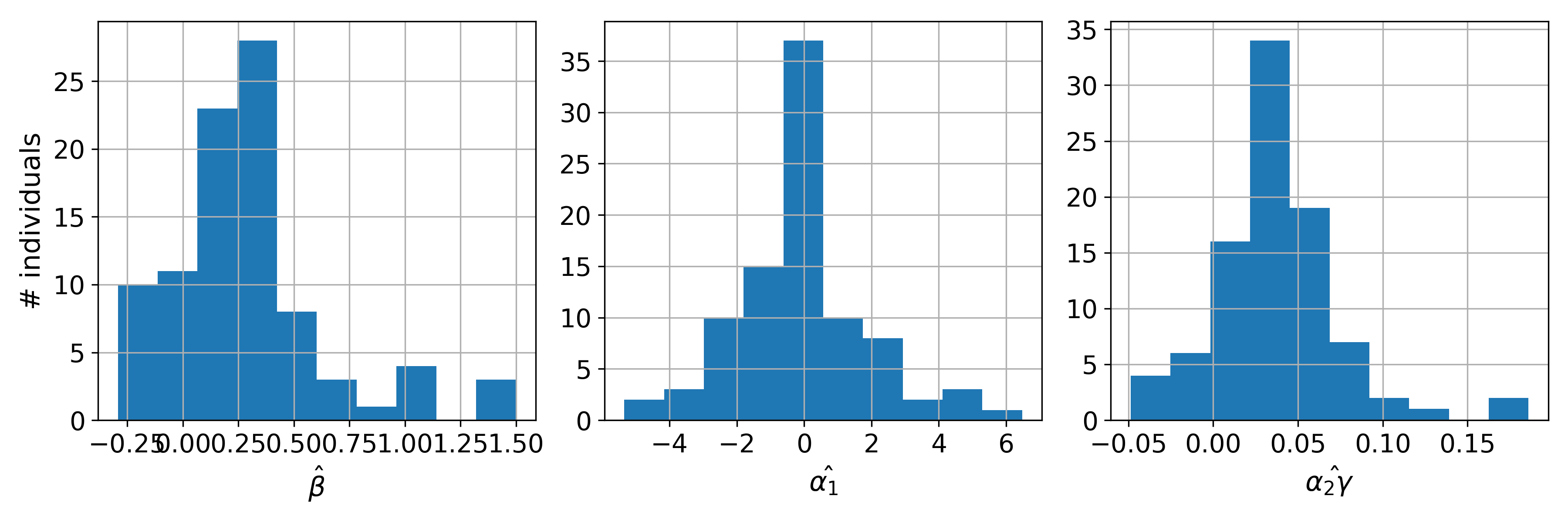}
    \caption{Individualized parameter estimation for split \#3 in ADNI. For visual clarity, only values between 5-95\%ile are plotted for each parameter.}
    \label{fig:perpat_paraes_adni}
\end{figure}

To identify the demographic variables $Z_0$ and their relation to parameters, $f(Z_0)$ (Eq. \ref{eq:params_demogs}), we followed a two step procedure. First, we estimated the parameters separately for each individual. Second, we performed statistical analysis to find associations between the estimated parameters and the demographic variables. The distribution of individualized parameter estimates are shown in Figure \ref{fig:perpat_paraes_adni}. During statistical analysis, we tested for association between each parameter and demographic variable pair. For discrete demographic variables, statistical association was tested with a Wilcoxon ranksum test. For continuous demographic variables, we fit a line with the demographic variable as the independent and the parameter as the dependent variable, and evaluated the p-value of the slope. Baseline age, education, APOE-$\epsilon$4 genotype, and gender were the demographic variables we considered. Parameter estimation and statistical analysis was done separately for each training split in the 5-fold cross-validation. The results of the statistical analysis are shown in Table \ref{tab:para_demog_stat}. Gender and genetic risk are the only demographic variables that are associated with some parameter in at least one split. Therefore, we chose those two to represent $Z_0$.

\begin{figure}
    \centering
    \includegraphics[width=0.95\linewidth]{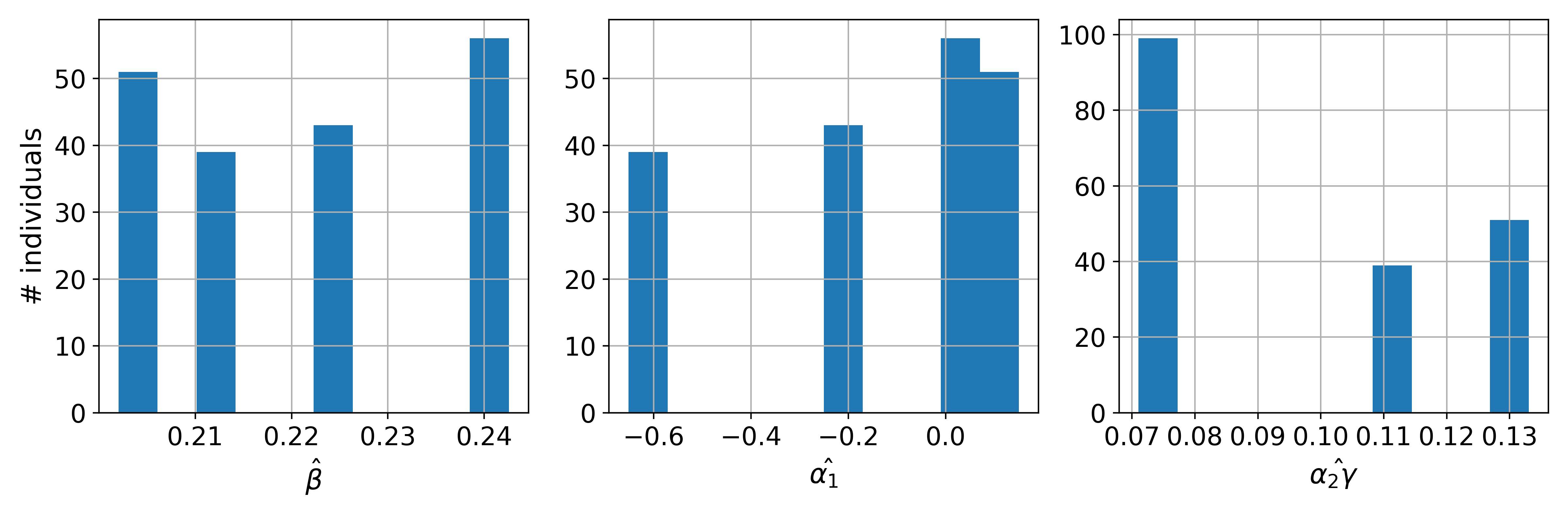}
    \caption{Group-wise parameter estimation for split \#3 in ADNI.}
    \label{fig:group_paraes_adni}
\end{figure}

\begin{table}[h]
    \centering
    \begin{tabular}{@{}ccccccc@{}}
        \toprule
        \textbf{Parameter}               & \textbf{Demographic ft.}     & \multicolumn{5}{c}{\textbf{$p$-value}}                                           \\ \midrule
                            &           & Split 0        & Split 1 & Split 2        & Split 3         & Split 4 \\ \midrule
        $\hat{\beta}$          & Age       & 0.472          & 0.669   & 0.796          & 0.79            & 0.557   \\
        $\hat{\beta}$          & Education & 0.589          & 0.61    & 0.591          & 0.378           & 0.947   \\
        $\hat{\beta}$          & Gender    & 0.381          & 0.802   & 0.708          & 0.766           & 0.879   \\
        $\hat{\beta}$          & \textbf{APOE-$\epsilon$4}   & \textbf{0.003} & 0.312   & 0.402          & \textbf{0.026}  & 0.076   \\ \hline
        $\hat{\alpha_1}$        & AGE       & 0.224          & 0.47    & 0.331          & 0.499           & 0.238   \\
        $\hat{\alpha_1}$        & Education & 0.227          & 0.915   & 0.331          & 0.114           & 0.951   \\
        $\hat{\alpha_1}$       & Gender    & 0.583          & 0.359   & 0.392          & 0.373           & 0.937   \\
        $\hat{\alpha_1}$        & APOE-$\epsilon$4   & 0.596          & 0.316   & 0.523          & 0.115           & 0.197   \\ \hline
        $\hat{\alpha_2\gamma}$ & Age       & 0.799          & 0.798   & 0.253          & 0.77            & 0.718   \\
        $\hat{\alpha_2\gamma}$ & Education & 0.904          & 0.133   & 0.821          & 0.244           & 0.079   \\
        $\hat{\alpha_2\gamma}$ & \textbf{Gender}    & 0.276          & 0.102   & \textbf{0.049} & \textbf{0.0005} & 0.056   \\
        $\hat{\alpha_2\gamma}$ & APOE-$\epsilon$4   & 0.884          & 0.492   & 0.357          & 0.234           & 0.915   \\ \bottomrule
    \end{tabular}
    \caption{Associations between individualized parameter estimates and demographic features. Cases where $p<0.05$ are highlighted.}
    \label{tab:para_demog_stat}
\end{table}
There are a large number of missing values in the ADNI data in our study. Estimating parameters based on groups mitigates the effect of missing values. Therefore, we estimated parameters for groups based on the gender and APOE-$\epsilon$4 status. This resulted in 4 groups. Parameter estimates for groups for ADNI data are shown in Figure \ref{fig:group_paraes_adni}.

\section{Data generation and pre-processing} \label{sec:appex_data}
\subsection{Synthetic data} 
We generated synthetic data of 200 samples (individuals) to validate the model's ability to model AD progression and predict long-term cognition trajectories. We set $|V|=2$ and generated longitudinal trajectories of $X(t), Y(t), D(t), \phi(t), I(t), \text{ and } C(t)$ for $K=10$ time points (+baseline). The baseline value of brain region size $X(0)$ and amyloid $D(0)$ were generated randomly; $X(0) \sim \mathcal{N}(\mu=[3.5, 3.4], \Sigma=[[0.49, 0.20],[0.20, 0.64]])$ and $D_{1}(0), D_{2}(0) \sim Uniform(0,0.2)$. Each individual also had two discrete demographic features, which were randomly sampled with equal probability from four and two possible values, respectively. This resulted in eight groups of individuals based on their demographic features. The parameters $\alpha_1, \alpha_2, \beta$ were computed from a predetermined linear combination of the demographic features. We set $\gamma=1$ to be the same for all the individuals and $C_{task}=10$. Finally, we set $I_{1}(t)= \min(Y_{max}X_{1}(t)/\gamma, C_{task})$ and $I_{2}(t)=\min(Y_{max}X_{2}(t)/\gamma, C_{task}-I_{1}(t))$ where $Y_{max}=2.5$. Based on the above and Eqs. \ref{eq:amyloid_spread}, \ref{eq:total_amyloid}, \ref{eq:cog_info}, \ref{eq:energy_info_health}, \ref{eq:health_atrophy}, we generated the longitudinal trajectories of $C(t), X(t), D(t), \phi(t), Y(t), I(t)$ for $t\in\{1,2,..,15\}$ and stored the last 11 time points to get trajectories that were heterogeneous.

\subsection{ADNI data} 
We used the ADNI dataset to evaluate the model on real-world data for Alzheimer's disease. For this study, we only included individuals who had (i) baseline (0$^{\text{th}}$ year) measurements of cognition, demographics, MRI, and florbetapir PET scans; (ii) longitudinal measurements of cognition; and (iii) at least 2 follow-up measurements (after baseline) that contain both PET and MRI scans along with cognitive assessment at those visits. Visits were not required to be successive, and only 10 years of assessments after baseline were retained for individuals with longer follow-ups. Note that cognitive assessments were retained for all available points upto and including year 10 irrespective of the availability of MRI/PET from those visits. These constraints were chosen to have sufficient measurements for (per-individual) parameter estimation. This resulted in data from 160 participants out of which 52 were cognitively normal (CN), 23 had significant memory concern (SMC), 58 had early mild cognitive impairment (EMCI), and 27 were diagnosed with late MCI (LMCI). 

Age, gender, education, and presence of APOE-$\epsilon$4 genotype were the demographic features. We considered a $|V|=2$ node graph $G_S$ with nodes representing the hippocampus (HC) and the prefrontal cortex (PFC) due to their importance in supporting cognition and their role in AD \cite{jack2010hypothetical, hillary2017injured, davis2008pasa}. AD pathology primarily targets the hippocampus, a region that plays an important role in memory and cognition, and propagates to other areas of the brain. PFC is involved in executive function in healthy individuals and shows increased activation in older adults and AD patients during cognitive tasks \cite{davis2008pasa}. Volume of the hippocampus (HC) and prefrontal cortex (PFC) were used to represent brain structure ($X(t)$). Raw values of hippocampal volume were divided by $2\times10^3$ to be close to 3. To account for the larger size of the PFC in the brain compared to the hippocampus, PFC volumes were normalized by the median ratio of PFC to hippocampus $\times 2\times10^3$. Median ratio was computed only on the training data. PET-scan derived SUVR values for PFC and hippocampus were used as a measure of A$\beta$ deposition ($\phi(t)$). Pre-processed files for MRI and PET available on LONI website were used. We used the score on the Mini Mental State Examination (MMSE) for as a measure of cognition. We scaled MMSE score by 3 such that 10 represented perfect cognition and 0 represented no cognition so as to match the choice of $C_{task}=10$. Although ADNI also contains functional MRI scans, we did not include them in our analysis, since very few individuals had that information along with the rest of the variables.

\begin{figure}
    \centering
    \includegraphics[width=0.4\linewidth]{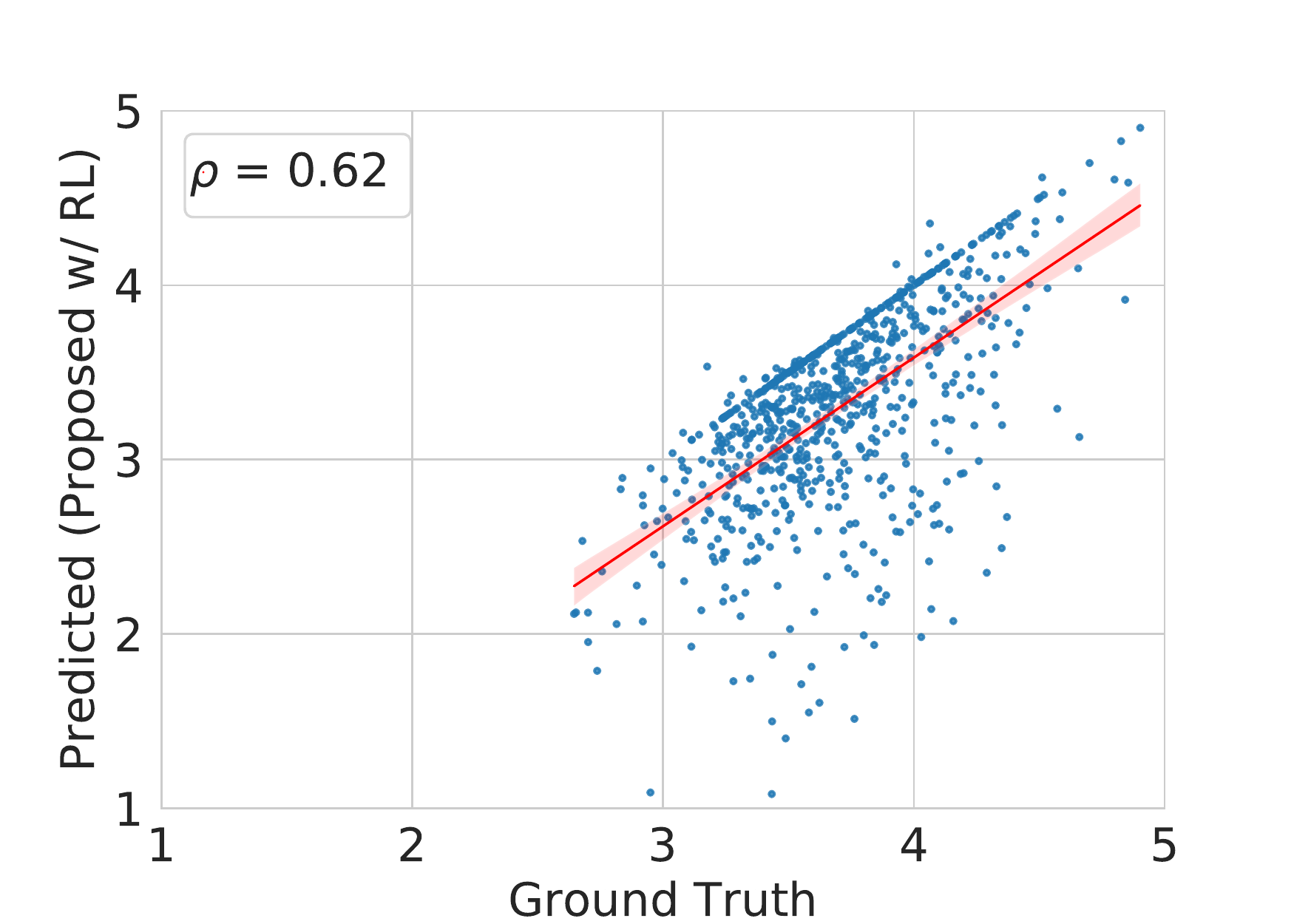}
    \caption{Predicted (proposed model with RL) vs ground truth size of PFC for ADNI data.}
    \label{fig:pfc_ground_predicted}
\end{figure}

\section{Model design choices and training}
\subsection{Effect of $\gamma$ on the model} \label{sec:appex_gamma_effect}
$\gamma$ parameter appears at two places in our model. In Eq. \ref{eq:health_atrophy} through $Y(t)$ and in the reward function Eq. \ref{eq:reward}. From the parameter estimators in Eq. \ref{eq:pares_alphas_gamma_withoutY}, it is clear that the effect of $\gamma$ on brain size occurs through the term $\alpha_2\gamma$ which can be estimated from the data. We can rewrite the reward function as:
\begin{align*}
    R(t) &= -\left[ \lambda(C_{task} - C(t)) + M(t) \right]\\
    &= -\left[ \lambda(C_{task} - \sum_{v\in V}I_v(t)) + \sum_{v \in V} \frac{\gamma I_v(t)}{X_v(t)} \right] \\
    &= -\gamma\left[ \frac{\lambda}{\gamma}(C_{task} - \sum_{v\in V}I_v(t)) + \sum_{v \in V} \frac{I_v(t)}{X_v(t)} \right]
\end{align*}
Since the optimization of the reward is performed over $I(t)$, $\gamma$ influences the optimization only through $\lambda/\gamma$. Therefore, we set $\gamma=1$ and varied $\lambda$ in our experiments.

For a similar reason, we modeled $M(t)=\sum_{v\in V}Y_v(t)$ instead of adding a proportionality constant. If there were a proportionality constant $\mu$ such that $M(t)=\mu\sum_{v\in V}Y_v(t)$, the effect of $\mu$ on the overall optimization would only occur through the term $\lambda/\mu\gamma$.

\subsection{Brain activity related degeneration} \label{sec:appex_brain_degen}
Although the effect of brain activity on neurodegeneration could be mediated through amyloid \cite{cirrito2005synaptic, hillary2017injured, jones2017tau}, we have included activity-based degeneration as a separate term in Eq. \ref{eq:health_atrophy}. This allows us to account for other pathways via which brain activity could lead to atrophy, e.g., oxidative stress and tauopathy. We also developed an alternate version of the proposed model that included an activity-related amyloid deposition term in Eq. \ref{eq:amyloid_spread} instead of the term in Eq. \ref{eq:health_atrophy}: 
\begin{align*}
    \frac{dD(t)}{dt} &= -\beta_1 H D(T) + \beta_2 Y(t) ; \text{ }
    \frac{dX_v(t)}{dt} = -\alpha D(t)
\end{align*}
The model with the above equations could not be trained on real data due to insufficient number of followups for parameter estimation of an individual. Due to the unavailability of $Y(t)$ and the second order dependence of $\frac{dD(t)}{dt}$ on $\phi(t)$, parameter estimators for $\beta_1, \beta_2$ required at least 4 measurements of $X(t), \phi(t), C(t)$ for estimation. Majority of the individuals (>100) in our data only had 3 measurements of $X(t), \phi(t)$. Nevertheless, we plan to appropriately include the effect of brain activity on degeneration via amyloid in future work.

\subsection{Modified relationship between $Y(t), X(t), I(t)$} \label{sec:appex_inv_sq}
\begin{figure}
    \centering
    \includegraphics[width=0.95\linewidth]{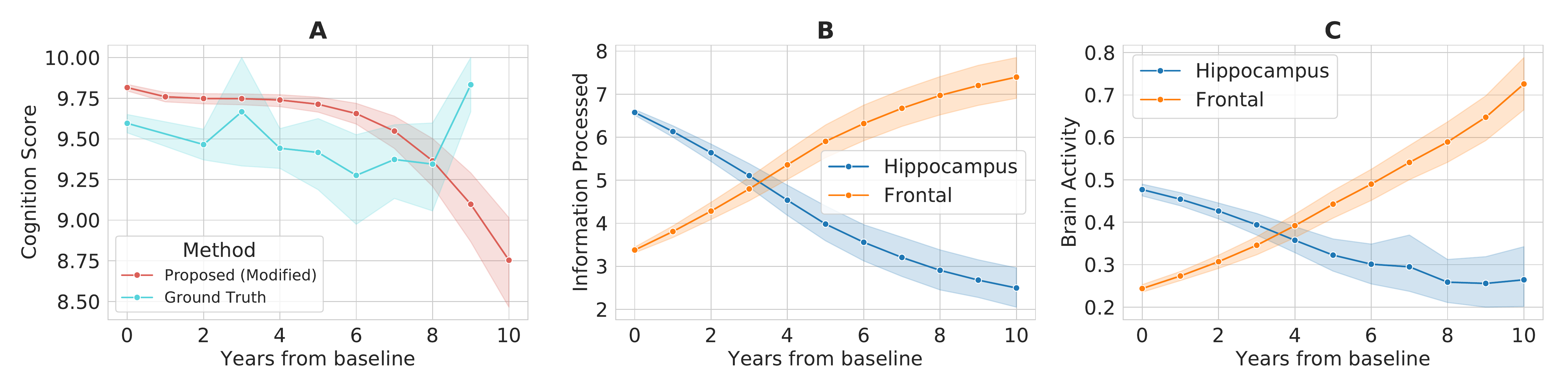}
    \caption{Modified model with inverse squared relationship between $Y(t)$ and $X(t)$ applied to ADNI data. (A) Cognition trajectories from ground truth and model. (B) Information processed in HC (Hippocampus) and PFC (Frontal) averaged across individuals. (C) Brain activity in HC and PFC averaged across individuals.}
    \label{fig:modified}
\end{figure}

In Eq \ref{eq:energy_info_health}, we used an inverse relationship between $Y(t)$ and $X(t)$ to capture the increase in activity for reduced region sizes. To assess the model's sensitivity to the form of the equations, we replaced Eq \ref{eq:energy_info_health} with a squared inverse relationship between $Y(t)$ and $X(t)$, as follows:
\begin{equation}
    Y_v(t) = \gamma \frac{I_v(t)}{X_v^{2}(t)} \quad \forall v \in V.
    \label{eq:energy_info_health_sq}
\end{equation}
For this modified model, we derived the parameter estimators following the procedure in Appendix \ref{sec:appex_param_es}; computed the parameters for each participant and group of participants; trained the RL agent with Eq \ref{eq:energy_info_health_sq} in the simulator instead of Eq \ref{eq:energy_info_health}; and evaluated cognition trajectory prediction performance and trends in information processing with disease progression. The modified model predicted cognitive trajectories with a performance comparable to other models presented in Table \ref{tab:cogpred} (MAE $= 0.539 (0.087)$ and MSE $= 0.786 (0.218)$) and demonstrated recovery/compensatory processes (Fig. \ref{fig:modified}).

\subsection{RL agent} \label{sec:appex_rltrain}
The agent determines the change in information processing in each region. Our simulator consisting of DEs (Eqs. \ref{eq:amyloid_spread}, \ref{eq:total_amyloid}, \ref{eq:cog_info}, \ref{eq:energy_info_health}, \ref{eq:metacost_energy}, \ref{eq:health_atrophy}, \ref{eq:reward}) allows us to use model-free RL with on-policy learning. During preliminary analysis (data not shown), we compared two of the most popular deep policy-gradient methods: TRPO \cite{schulman2015trust} and PPO \cite{schulman2017proximal}. For both methods, we set the discount factor for generalized advantage estimates to be 0.97. For PPO, we clipped the likelihood ratio at 0.2 and used a minibatch size of 10. For TRPO, we imposed a constrain of 0.01 on the KL-divergence. We used the default values for the remaining hyperparameters. We compared the reward achieved by agents trained using TRPO and PPO and observed that TRPO outperformed PPO, for both synthetic and ADNI data (data now shown).

In our work, the policy agent is parameterized by a multilayer, feedforward neural network. We implemented the environment and the agent's interaction using OpenAI's Gym framework \cite{brockman2016openai}. We used a stochastic two-hidden-layer Gaussian MLP with 32 neurons as the policy network, which we trained using the Garage Framework \cite{garage} (available under MIT License). During each epoch, the policy sampled 1000 trajectories from the simulator. An individual's trajectory consists of 11 time points including baseline. To stabilize learning, we imposed a minimum constraint of `-2000' on the reward and constrained the continuous action space in the range [-2,2]. The constraints on the action space were also motivated by the change in MMSE scores between subsequent years in ADNI data, $>95\%$ of which lie in the range [-2, 2].
To constrain the agent from assigning $C(t)>10$ to an individual, we incorporated a penalty factor in the reward function, based on the mismatch between $C_{task}$ and $C(t)$: $100^{[\max(C(t)-C_{task},0)]}$. Therefore, effectively, $R(t) = -[\lambda |C_{task}-C(t)|\times100^{[\max(C(t)-C_{task},0)]} + M(t)]$. The model was trained on an internal multi-node compute cluster with two 20-core IBM POWER9 CPUs at 2.4GHz and 256 GB RAM.

The model requires $D(t)$ for the simulation although only $\phi(0)$ is available from baseline data. $D(1)$ is computed from $\phi(0)$ separately for each individual using the relationship between them provided in \cite{raj2015network}: 
\begin{align}
    \frac{d\phi(t_{po})}{dt_{po}} &= \beta \Tilde{H}(\beta t_{po})\phi(t_{po}), \label{eq:phi_to_D}\\
    \Tilde{H}(\beta t_{po}) &= U diag\left(\left\{
    	\begin{array}{ll}
    		\frac{1}{\beta t_{po}}  & \mbox{if } j=0 \\
    		\frac{\nu_{j}e^{-\nu_{j}\beta t_{po}}}{1-e^{-\nu_{j}\beta t_{po}}} & \mbox{if } j > 0
    	\end{array}
    \right\} \right) U^{T}, \nonumber
\end{align}
where $\nu_{j}$ are the eigenvalues and $U$ contains the corresponding eigenvectors of $H$, and $t_{po}$ denotes the time post onset of amyloid deposition. We approximate Eq. \ref{eq:phi_to_D} to calculate $D(1) = \frac{\Delta \phi(t_{po})}{\Delta t_{po}} = \phi(t_{po}+1) - \phi(t_{po})$, and set $\phi(t_{po}) = \phi(0)$ (amyloid deposition at baseline) and $t_{po}=$ individual's age at baseline $- 50$.

We follow a two step procedure to determine $I(0)$ for an individual. First, we determine the best $I(0)$ for the entire population by using grid search as described in Section \ref{sec:appex_grid}. Second, we fine-tuned the $I^{i}(0)$ for an individual $i$ using the policy agent based on (i) their brain regions size $X^{i}(0)$, and (ii) the population $I(0)$ found from grid search. We observed that this two step procedure of selecting a population based $I(0)$ and then fine-tuning it for an individual allowed us to account for the variability in baseline cognition across individuals.

\subsection{Grid search for choosing $\lambda$ and $I(0)$} \label{sec:appex_grid}
$\lambda$ and $I(0)$ must be specified for training the model and for predicting disease progression for an individual. We performed a grid search for $\lambda$ and $I(0)$ and chose the optimal value based on the average validation set error from cross validation. Grid search was performed independently for the synthetic data and for the ADNI data. We varied $\lambda \in \{2^{-1}, 2^{0}, 2^{1}, 2^{2}, 2^{3}\}$ for the synthetic data and $\lambda \in \{2^{-2},2^{-1},2^{0},2^{1}\}$ for the ADNI data based on preliminary analysis. $I(0)$ was chosen from the set $\{[10,0], [9,1], [8,2], [7,3], [6,4]\}$ for both datasets, with the representation $I(0) = [I_{HC}(0), I_{PFC}(0)]$ for ADNI data. Based on the validation MSE (Table \ref{tab:gridsearch}), we chose $\lambda=2.0$ for synthetic data and $\lambda=1.0$ for ADNI data.  The optimal validation set error was achieved by $I(0)=[7,3]$ for ADNI and $I(0)=[9,1]$ for synthetic data (Table \ref{tab:gridsearch}). The value of $\lambda$ and $I(0)$ identified during training was used while testing.

\begin{table}[h]
    \centering
        \resizebox{\textwidth}{!}{%
            \begin{tabular}{|c|rrrr|rrrrr|}
            \hline
            \multicolumn{1}{|l|}{\textbf{MAE}} & \multicolumn{4}{c|}{\textbf{ADNI}} & \multicolumn{5}{c|}{\textbf{Synthetic}} \\ \cline{2-10} 
            \multicolumn{1}{|l|}{} & \multicolumn{1}{c}{\textbf{0.25}} & \multicolumn{1}{c}{\textbf{0.5}} & \multicolumn{1}{c}{\textbf{1}} & \multicolumn{1}{c|}{\textbf{2}} & \multicolumn{1}{c}{\textbf{0.5}} & \multicolumn{1}{c}{\textbf{1}} & \multicolumn{1}{c}{\textbf{2}} & \multicolumn{1}{c}{\textbf{4}} & \multicolumn{1}{c|}{\textbf{8}} \\ \hline
            \textbf{6,4} & 7.730 (0.261) & 0.705 (0.180) & 0.499 (0.059) & 0.520 (0.093) & 2.230 (0.225) & 0.806 (0.069) & 0.711 (0.032) & 0.699 (0.116) & 0.754 (0.204) \\ \hline
            \textbf{7,3} & 7.457 (0.255) & 0.724 (0.191) & \textbf{0.485 (0.087)} & 0.492 (0.091) & 2.292 (0.249) & 0.741 (0.115) & 0.766 (0.215) & 0.713 (0.084) & 0.700 (0.050) \\ \hline
            \textbf{8,2} & 7.264 (0.274) & 0.732 (0.174) & 0.516 (0.072) & 0.495 (0.071) & 2.296 (0.094) & 0.764 (0.070) & 0.698 (0.112) & 0.739 (0.138) & 0.690 (0.163) \\ \hline
            \textbf{9,1} & 6.805 (0.283) & 0.824 (0.178) & 0.529 (0.076) & 0.522 (0.093) & 2.166 (0.221) & 0.892 (0.206) & \textbf{0.670 (0.053)} & 0.781 (0.230) & 0.808 (0.228) \\ \hline
            \textbf{10,0} & 6.348 (0.227) & 0.822 (0.346) & 0.527 (0.075) & 0.497 (0.099) & 2.057 (0.188) & 0.834 (0.130) & 0.904 (0.376) & 0.885 (0.321) & 0.891 (0.168) \\ \hline
            \end{tabular}
        }
    \caption{Average validation MAE for different values of $\lambda$ and $I(0)$ for ADNI and synthetic data. Standard deviation is provided in parentheses. The lowest MAE for each dataset is highlighted.}
    \label{tab:gridsearch}
\end{table}

\section{Benchmark and control models} \label{sec:appex_bench}
\subsection{minimalRNN} Minimal recurrent neural network is a state-of-the-art model to predict cognition trajectories \cite{NGUYEN2020117203}. The model predicts AD progression up to 6 years into the future using data from one (baseline) or more time points. We used the open-source implementation of minimalRNN (\href{https://github.com/ThomasYeoLab/CBIG/tree/master/stable\_projects/predict\_phenotypes/Nguyen2020\_RNNAD}{GitHub link}; available under MIT License). To maintain consistency of experiments across different models, only data at year 0 was input into the model. "Model filling" was used to infer values of brain region size ($X(t)$), amyloid pathology ($\phi(t)$), and age in subsequent years (see \cite{NGUYEN2020117203} for details). With the inferred $X(t)$, $\phi(t)$, and age along with other demographics (genetics and gender; $Z_0$), the minimalRNN model predicted cognition ($C(t)$) at each year. In this study, minimalRNN was trained for $100$ epochs with a learning rate of $5\times 10^{-4}$ and batch size of $128$. The model is optimized by an Adam optimizer with weight decay set to $5\times 10^{-7}$.

\subsection{SVR} 
Kernel support vector regression has been used as a benchmark model for cognition trajectory prediction \cite{NGUYEN2020117203}. It is a suitable learning based model in scenarios where the available data is limited. In our experimental settings, baseline brain region size ($X(0)$); amyloid pathology ($\phi(0)$); and demographics gender, genetic risk ($Z_0$), and age were input to the SVR model. SVR was implemented with the RBF kernel (with hyper-parameters: $C=1$, $epsilon=0.1$ and $gamma=0.01$).

\subsection{Proposed model without RL} \label{sec:appex_worl} 
\begin{figure}
    \centering
    \includegraphics[width=0.8\linewidth]{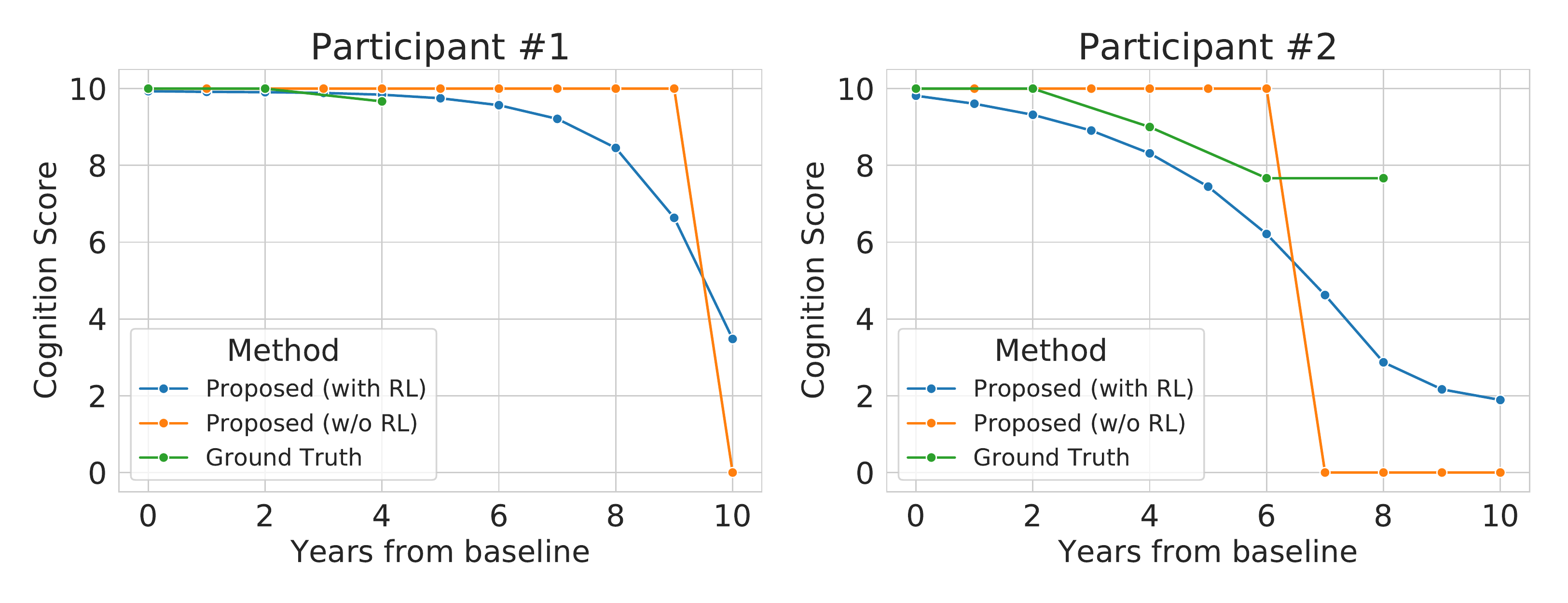}
    \caption{Cognition trajectories for two individuals. Ground truth trajectory and predicted trajectories (from proposed model with RL and proposed model without RL) are shown.}
    \label{fig:sample_curves}
\end{figure}

To assess the value of RL in our framework, we implemented our DE-based model and optimized $R(t)$ (Sec. \ref{sec:appex_rltrain}) at each time point using standard optimization instead of RL. Optimization of $R(t)$ was done independently at each time point. For each time point $t$, we performed a grid search by varying $I_v(t) \in \{0, 0.1, 0.2,..., 10.4\} \quad \forall v \in V$ and chose the value of $I(t)$ that maximized $R(t)$. $\lambda$ was the same as in the best performing proposed model reported in Table \ref{tab:cogpred}. The proposed model w/o RL as described above typically resulted in $I(t) \in \{[10,0], [0,10], [0,0]\} \quad \forall t$ (Fig. \ref{fig:sample_curves}).

\subsection{Control model for recovery} \label{sec:appex_control_mod}
\begin{figure}[h]
    \centering
    \includegraphics[width=\linewidth]{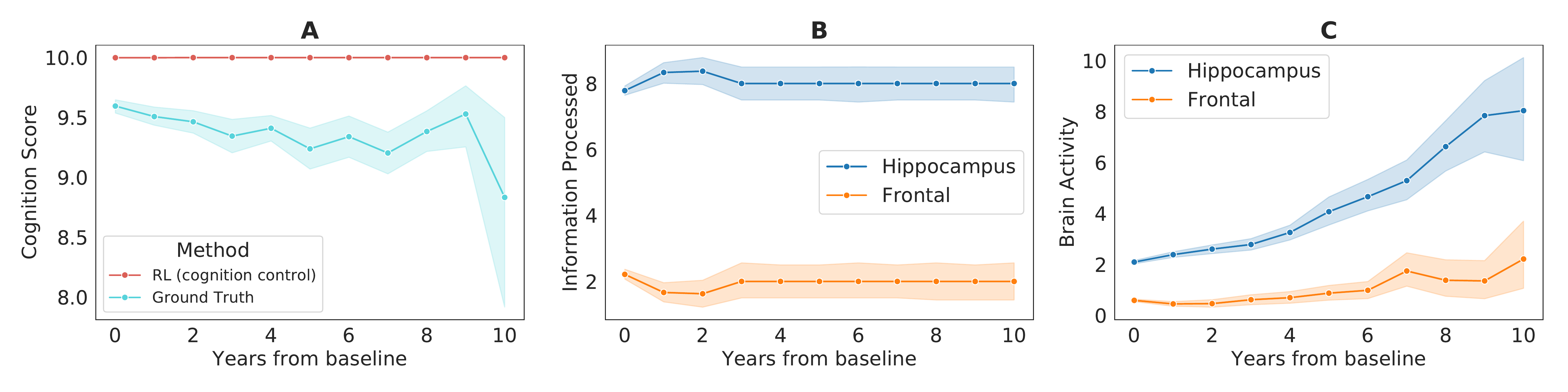}
    \caption{Control model with reward $R'(t)$ for ADNI. (A) Cognition trajectories from ground truth and model. (B) Information processed in HC (Hippocampus) and PFC (Frontal) averaged across individuals. (C) Brain activity in HC and PFC averaged across individuals.}
    \label{fig:controlreward_cogn}
\end{figure}

\begin{figure}[h]
    \includegraphics[width=\linewidth]{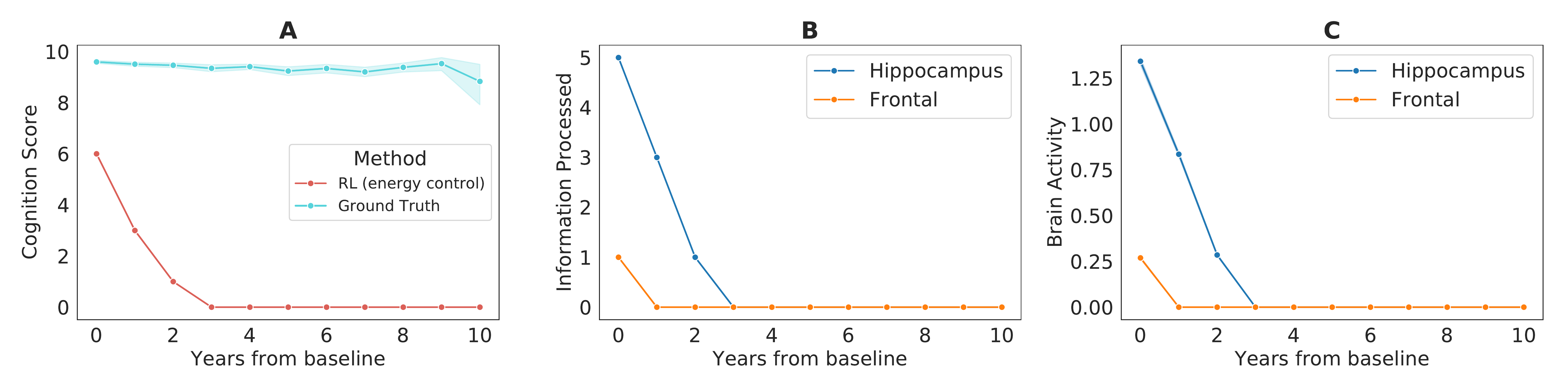}
    \caption{Control model with reward $R''(t)$ for ADNI. (A) Cognition trajectories from ground truth and model. (B) Information processed in HC (Hippocampus) and PFC (Frontal) averaged across individuals. (C) Brain activity in HC and PFC averaged across individuals.}
    \label{fig:controlreward_energy}
\end{figure}

We implemented control models to evaluate the effect of the reward function on the recovery like behavior observed in Sec. \ref{sec:results}. Specifically, we hypothesized that the shift in information processing observed in the proposed model was due to the form of the reward function $R(t)$ which consisted of a (i) cognitive mismatch term, and an (ii) energetic cost of cognition term. To test this hypothesis, we implemented two variants of the proposed model as control models with modified reward functions $R'(t) = -(C_{task}-C(t))$ and $R''(t) = M(t)$, respectively. The modified reward functions only consisted of one term each from $R(t)$. Note that, unlike $R(t)$ (Eq. \ref{eq:reward}), the modified reward functions did not need a $\lambda$ parameter. I(0) for the control model was the same as in the best model from Table \ref{tab:cogpred}. Information processing plots for the control models trained on ADNI data are shown in Figs. \ref{fig:controlreward_cogn}, \ref{fig:controlreward_energy}. 

The control model with reward $R'(t)$ roughly maintains the initial information processing distribution over regions, $I(0)$, throughout the entire 10 years (Fig. \ref{fig:controlreward_cogn}). Maintaining $I_{1}(t)$ at a high value leads to increased degeneration of the HC, which consequently increases $M(t)$. Since increased $M(t)$ does not penalize the reward, the control model approximately maintains $I(0)$ throughout. This suggests that $M(t)$ plays a role in demonstrating recovery in the proposed model. 

The control model with reward $R''(t)$ pushes information processing in both regions to 0, i.e., $I(t)=[0,0], \; t \geq 1$ (Fig. \ref{fig:controlreward_energy}). As expected, this model minimizes the energetic cost, i.e., $M(t)=0$. This results in $C(t)=0, \; t \geq 3$.

In summary, neither control model with the modified reward functions demonstrated recovery. Moreover, they highlighted the value of the two competing terms of (i) cognitive mismatch, and (ii) energetic cost in the reward function in demonstrating recovery in the proposed model. 

%% file: main.bbl
\begin{thebibliography}{10}

\bibitem{batsch2015world}
Nicole~L Batsch and Mary~S Mittelman.
\newblock World alzheimer report 2012.
\newblock {\em Overcoming the Stigma of Dementia. Alzheimer’s Disease
  International (ADI), London; 2012. Accessed May}, 5, 2015.

\bibitem{breijyeh2020comprehensive}
Zeinab Breijyeh and Rafik Karaman.
\newblock Comprehensive review on alzheimer’s disease: Causes and treatment.
\newblock {\em Molecules}, 25(24):5789, 2020.

\bibitem{raj2012network}
Ashish Raj, Amy Kuceyeski, and Michael Weiner.
\newblock A network diffusion model of disease progression in dementia.
\newblock {\em Neuron}, 73(6):1204--1215, 2012.

\bibitem{weickenmeier2018multiphysics}
Johannes Weickenmeier, Ellen Kuhl, and Alain Goriely.
\newblock Multiphysics of prionlike diseases: Progression and atrophy.
\newblock {\em Physical review letters}, 121(15):158101, 2018.

\bibitem{raj2015network}
Ashish Raj, Eve LoCastro, Amy Kuceyeski, Duygu Tosun, Norman Relkin, Michael
  Weiner, Alzheimer’s Disease Neuroimaging~Initiative (ADNI, et~al.
\newblock Network diffusion model of progression predicts longitudinal patterns
  of atrophy and metabolism in alzheimer’s disease.
\newblock {\em Cell reports}, 10(3):359--369, 2015.

\bibitem{conrado2020challenges}
Daniela~J Conrado, Sridhar Duvvuri, Hugo Geerts, Jackson Burton, Carla
  Biesdorf, Malidi Ahamadi, Sreeraj Macha, Gregory Hather, Juan
  Francisco~Morales, Jagdeep Podichetty, et~al.
\newblock Challenges in alzheimer's disease drug discovery and development: The
  role of modeling, simulation, and open data.
\newblock {\em Clinical Pharmacology \& Therapeutics}, 107(4):796--805, 2020.

\bibitem{bassett2009cognitive}
Danielle~S Bassett, Edward~T Bullmore, Andreas Meyer-Lindenberg, Jos{\'e}~A
  Apud, Daniel~R Weinberger, and Richard Coppola.
\newblock Cognitive fitness of cost-efficient brain functional networks.
\newblock {\em Proceedings of the National Academy of Sciences},
  106(28):11747--11752, 2009.

\bibitem{ito2017cognitive}
Takuya Ito, Kaustubh~R Kulkarni, Douglas~H Schultz, Ravi~D Mill, Richard~H
  Chen, Levi~I Solomyak, and Michael~W Cole.
\newblock Cognitive task information is transferred between brain regions via
  resting-state network topology.
\newblock {\em Nature communications}, 8(1):1--14, 2017.

\bibitem{christie2015cognitive}
Scott~Thomas Christie and Paul Schrater.
\newblock Cognitive cost as dynamic allocation of energetic resources.
\newblock {\em Frontiers in neuroscience}, 9:289, 2015.

\bibitem{NGUYEN2020117203}
Minh Nguyen, Tong He, Lijun An, Daniel~C. Alexander, Jiashi Feng, and
  B.T.~Thomas Yeo.
\newblock Predicting alzheimer's disease progression using deep recurrent
  neural networks.
\newblock {\em NeuroImage}, 222:117203, 2020.

\bibitem{hillary2017injured}
Frank~G Hillary and Jordan~H Grafman.
\newblock Injured brains and adaptive networks: the benefits and costs of
  hyperconnectivity.
\newblock {\em Trends in cognitive sciences}, 21(5):385--401, 2017.

\bibitem{jack2010hypothetical}
Clifford~R Jack~Jr, David~S Knopman, William~J Jagust, Leslie~M Shaw, Paul~S
  Aisen, Michael~W Weiner, Ronald~C Petersen, and John~Q Trojanowski.
\newblock Hypothetical model of dynamic biomarkers of the alzheimer's
  pathological cascade.
\newblock {\em The Lancet Neurology}, 9(1):119--128, 2010.

\bibitem{davis2008pasa}
Simon~W Davis, Nancy~A Dennis, Sander~M Daselaar, Mathias~S Fleck, and Roberto
  Cabeza.
\newblock Que pasa? the posterior--anterior shift in aging.
\newblock {\em Cerebral cortex}, 18(5):1201--1209, 2008.

\bibitem{dennis2014functional}
Emily~L Dennis and Paul~M Thompson.
\newblock Functional brain connectivity using fmri in aging and alzheimer’s
  disease.
\newblock {\em Neuropsychology review}, 24(1):49--62, 2014.

\bibitem{jones2017tau}
David~T Jones, Jonathan Graff-Radford, Val~J Lowe, Heather~J Wiste, Jeffrey~L
  Gunter, Matthew~L Senjem, Hugo Botha, Kejal Kantarci, Bradley~F Boeve,
  David~S Knopman, et~al.
\newblock Tau, amyloid, and cascading network failure across the alzheimer's
  disease spectrum.
\newblock {\em Cortex}, 97:143--159, 2017.

\bibitem{braak2000vulnerability}
Heiko Braak, Kelly del Tredici, Christian Schultz, and Eva Braak.
\newblock Vulnerability of select neuronal types to alzheimer's disease.
\newblock {\em Annals of the New York Academy of Sciences}, 924(1):53--61,
  2000.

\bibitem{macaulay2020predictors}
Rebecca~K MacAulay, Amy Halpin, Alex~S Cohen, Matthew Calamia, Angelica Boeve,
  Le~Zhang, Robert~M Brouillette, Heather~C Foil, Annadora Bruce-Keller, and
  Jeffrey~N Keller.
\newblock Predictors of heterogeneity in cognitive function: Apoe-e4, sex,
  education, depression, and vascular risk.
\newblock {\em Archives of Clinical Neuropsychology}, 35(6):660--670, 2020.

\bibitem{honey2007network}
Christopher~J Honey, Rolf K{\"o}tter, Michael Breakspear, and Olaf Sporns.
\newblock Network structure of cerebral cortex shapes functional connectivity
  on multiple time scales.
\newblock {\em Proceedings of the National Academy of Sciences},
  104(24):10240--10245, 2007.

\bibitem{honey2009predicting}
Christopher~J Honey, Olaf Sporns, Leila Cammoun, Xavier Gigandet, Jean-Philippe
  Thiran, Reto Meuli, and Patric Hagmann.
\newblock Predicting human resting-state functional connectivity from
  structural connectivity.
\newblock {\em Proceedings of the National Academy of Sciences},
  106(6):2035--2040, 2009.

\bibitem{stern2012task}
Yaakov Stern, Brian~C Rakitin, Christian Habeck, Yunglin Gazes, Jason
  Steffener, Arjun Kumar, and Aaron Reuben.
\newblock Task difficulty modulates young--old differences in network
  expression.
\newblock {\em Brain research}, 1435:130--145, 2012.

\bibitem{niven2007fly}
Jeremy~E Niven, John~C Anderson, and Simon~B Laughlin.
\newblock Fly photoreceptors demonstrate energy-information trade-offs in
  neural coding.
\newblock {\em PLoS Biol}, 5(4):e116, 2007.

\bibitem{attwell2001energy}
David Attwell and Simon~B Laughlin.
\newblock An energy budget for signaling in the grey matter of the brain.
\newblock {\em Journal of Cerebral Blood Flow \& Metabolism},
  21(10):1133--1145, 2001.

\bibitem{schulman2015trust}
John Schulman, Sergey Levine, Pieter Abbeel, Michael Jordan, and Philipp
  Moritz.
\newblock Trust region policy optimization.
\newblock In {\em International conference on machine learning}, pages
  1889--1897, 2015.

\bibitem{brockman2016openai}
Greg Brockman, Vicki Cheung, Ludwig Pettersson, Jonas Schneider, John Schulman,
  Jie Tang, and Wojciech Zaremba.
\newblock Openai gym.
\newblock {\em arXiv preprint arXiv:1606.01540}, 2016.

\bibitem{weiner2015impact}
Michael~W Weiner, Dallas~P Veitch, Paul~S Aisen, Laurel~A Beckett, Nigel~J
  Cairns, Jesse Cedarbaum, Michael~C Donohue, Robert~C Green, Danielle Harvey,
  Clifford~R Jack~Jr, et~al.
\newblock Impact of the alzheimer's disease neuroimaging initiative, 2004 to
  2014.
\newblock {\em Alzheimer's \& Dementia}, 11(7):865--884, 2015.

\bibitem{raket2020statistical}
Lars~Lau Raket, Alzheimer's Disease~Neuroimaging Initiative, et~al.
\newblock Statistical disease progression modeling in alzheimer disease.
\newblock {\em Frontiers in Big Data}, 3, 2020.

\bibitem{ashraf2015cortical}
A~Ashraf, Z~Fan, DJ~Brooks, and P~Edison.
\newblock Cortical hypermetabolism in mci subjects: a compensatory mechanism?
\newblock {\em European journal of nuclear medicine and molecular imaging},
  42(3):447--458, 2015.

\bibitem{fonteijn2012event}
Hubert~M Fonteijn, Marc Modat, Matthew~J Clarkson, Josephine Barnes, Manja
  Lehmann, Nicola~Z Hobbs, Rachael~I Scahill, Sarah~J Tabrizi, Sebastien
  Ourselin, Nick~C Fox, et~al.
\newblock An event-based model for disease progression and its application in
  familial alzheimer's disease and huntington's disease.
\newblock {\em NeuroImage}, 60(3):1880--1889, 2012.

\bibitem{oxtoby2018data}
Neil~P Oxtoby, Alexandra~L Young, David~M Cash, Tammie~LS Benzinger, Anne~M
  Fagan, John~C Morris, Randall~J Bateman, Nick~C Fox, Jonathan~M Schott, and
  Daniel~C Alexander.
\newblock Data-driven models of dominantly-inherited alzheimer’s disease
  progression.
\newblock {\em Brain}, 141(5):1529--1544, 2018.

\bibitem{fruehwirt2018bayesian}
Wolfgang Fruehwirt, Adam~D Cobb, Martin Mairhofer, Leonard Weydemann, Heinrich
  Garn, Reinhold Schmidt, Thomas Benke, Peter Dal-Bianco, Gerhard Ransmayr,
  Markus Waser, et~al.
\newblock Bayesian deep neural networks for low-cost neurophysiological markers
  of alzheimer's disease severity.
\newblock {\em arXiv preprint arXiv:1812.04994}, 2018.

\bibitem{lin2018convolutional}
Weiming Lin, Tong Tong, Qinquan Gao, Di~Guo, Xiaofeng Du, Yonggui Yang, Gang
  Guo, Min Xiao, Min Du, Xiaobo Qu, et~al.
\newblock Convolutional neural networks-based mri image analysis for the
  alzheimer’s disease prediction from mild cognitive impairment.
\newblock {\em Frontiers in neuroscience}, 12:777, 2018.

\bibitem{tabarestani2018profile}
Solale Tabarestani, Maryamossadat Aghili, Mehdi Shojaie, Christian Freytes, and
  Malek Adjouadi.
\newblock Profile-specific regression model for progression prediction of
  alzheimer's disease using longitudinal data.
\newblock In {\em 2018 17th IEEE International Conference on Machine Learning
  and Applications (ICMLA)}, pages 1353--1357. IEEE, 2018.

\bibitem{saboo2020predicting}
Krishnakant Saboo, Chang Hu, Yogatheesan Varatharajah, Prashanthi Vemuri, and
  Ravishankar Iyer.
\newblock Predicting longitudinal cognitive scores using baseline imaging and
  clinical variables.
\newblock In {\em 2020 IEEE 17th International Symposium on Biomedical Imaging
  (ISBI)}, pages 1326--1330. IEEE, 2020.

\bibitem{marinescu2019tadpole}
R{\u{a}}zvan~V Marinescu, Neil~P Oxtoby, Alexandra~L Young, Esther~E Bron,
  Arthur~W Toga, Michael~W Weiner, Frederik Barkhof, Nick~C Fox, Polina
  Golland, Stefan Klein, et~al.
\newblock Tadpole challenge: Accurate alzheimer’s disease prediction through
  crowdsourced forecasting of future data.
\newblock In {\em International Workshop on PRedictive Intelligence In
  MEdicine}, pages 1--10. Springer, 2019.

\bibitem{jack2011evidence}
Clifford~R Jack, Prashanthi Vemuri, Heather~J Wiste, Stephen~D Weigand, Paul~S
  Aisen, John~Q Trojanowski, Leslie~M Shaw, Matthew~A Bernstein, Ronald~C
  Petersen, Michael~W Weiner, et~al.
\newblock Evidence for ordering of alzheimer disease biomarkers.
\newblock {\em Archives of neurology}, 68(12):1526--1535, 2011.

\bibitem{anderson2017so}
Roy~M Anderson, Christoforos Hadjichrysanthou, Stephanie Evans, and Mei~Mei
  Wong.
\newblock Why do so many clinical trials of therapies for alzheimer's disease
  fail?
\newblock {\em The Lancet}, 390(10110):2327--2329, 2017.

\bibitem{li2016simulating}
Wei Li, Miao Wang, Wenzhen Zhu, Yuanyuan Qin, Yue Huang, and Xi~Chen.
\newblock Simulating the evolution of functional brain networks in
  alzheimer’s disease: exploring disease dynamics from the perspective of
  global activity.
\newblock {\em Scientific reports}, 6:34156, 2016.

\bibitem{vertes2012simple}
Petra~E V{\'e}rtes, Aaron~F Alexander-Bloch, Nitin Gogtay, Jay~N Giedd,
  Judith~L Rapoport, and Edward~T Bullmore.
\newblock Simple models of human brain functional networks.
\newblock {\em Proceedings of the National Academy of Sciences},
  109(15):5868--5873, 2012.

\bibitem{frassle2018generative}
Stefan Fr{\"a}ssle, Ekaterina~I Lomakina, Lars Kasper, Zina~M Manjaly, Alex
  Leff, Klaas~P Pruessmann, Joachim~M Buhmann, and Klaas~E Stephan.
\newblock A generative model of whole-brain effective connectivity.
\newblock {\em Neuroimage}, 179:505--529, 2018.

\bibitem{pineau2009treating}
Joelle Pineau, Arthur Guez, Robert Vincent, Gabriella Panuccio, and Massimo
  Avoli.
\newblock Treating epilepsy via adaptive neurostimulation: a reinforcement
  learning approach.
\newblock {\em International journal of neural systems}, 19(04):227--240, 2009.

\bibitem{krylov2020reinforcement}
Dmitrii Krylov, Remi Tachet, Romain Laroche, Michael Rosenblum, and Dmitry~V
  Dylov.
\newblock Reinforcement learning framework for deep brain stimulation study.
\newblock {\em arXiv preprint arXiv:2002.10948}, 2020.

\bibitem{venuto2016review}
Charles~S Venuto, Nicholas~B Potter, E~Ray~Dorsey, and Karl Kieburtz.
\newblock A review of disease progression models of parkinson's disease and
  applications in clinical trials.
\newblock {\em Movement Disorders}, 31(7):947--956, 2016.

\bibitem{orlowski2013modelling}
Piotr Orlowski, David O'Neill, Vicente Grau, Yiannis Ventikos, and Stephen
  Payne.
\newblock Modelling of the physiological response of the brain to ischaemic
  stroke.
\newblock {\em Interface focus}, 3(2):20120079, 2013.

\bibitem{cirrito2005synaptic}
John~R Cirrito, Kelvin~A Yamada, Mary~Beth Finn, Robert~S Sloviter, Kelly~R
  Bales, Patrick~C May, Darryle~D Schoepp, Steven~M Paul, Steven Mennerick, and
  David~M Holtzman.
\newblock Synaptic activity regulates interstitial fluid amyloid-$\beta$ levels
  in vivo.
\newblock {\em Neuron}, 48(6):913--922, 2005.

\bibitem{schulman2017proximal}
John Schulman, Filip Wolski, Prafulla Dhariwal, Alec Radford, and Oleg Klimov.
\newblock Proximal policy optimization algorithms.
\newblock {\em arXiv preprint arXiv:1707.06347}, 2017.

\bibitem{garage}
The garage contributors.
\newblock Garage: A toolkit for reproducible reinforcement learning research.
\newblock \url{https://github.com/rlworkgroup/garage}, 2019.

\end{thebibliography}
